\documentclass{article}

\usepackage[numbers,compress]{natbib}
\usepackage[final]{nips_2017}

\usepackage[utf8]{inputenc} % allow utf-8 input
\usepackage[T1]{fontenc}    % use 8-bit T1 fonts
\usepackage{hyperref}       % hyperlinks
\usepackage{url}            % simple URL typesetting
\usepackage{booktabs}       % professional-quality tables
\usepackage{amsfonts}       % blackboard math symbols
\usepackage{nicefrac}       % compact symbols for 1/2, etc.
\usepackage{microtype}      % microtypography

\usepackage{times}
\usepackage{epsfig}
\usepackage{graphicx}
\usepackage{amsmath}
\usepackage{amssymb}
\usepackage{amsfonts}
\usepackage{tabularx}
\usepackage{booktabs}
\usepackage{comment}
\usepackage{soul}
\usepackage{xcolor}

\newcommand{\beas}{\begin{eqnarray*}}
	\newcommand{\eeas}{\end{eqnarray*}}
\newcommand{\bea}{\begin{eqnarray}}
\newcommand{\eea}{\end{eqnarray}}
\newcommand{\bes}{\begin{equation*}}
\newcommand{\ees}{\end{equation*}}
\newcommand{\be}{\begin{equation}}
\newcommand{\ee}{\end{equation}}

\newcommand{\cY}{{\cal Y}}

\newcommand{\twocmidrule}{\cmidrule(lr){1-1}\cmidrule(lr){2-2}}

\makeatletter
\usepackage{xspace}
\def\@onedot{\ifx\@let@token.\else.\null\fi\xspace}
\DeclareRobustCommand\onedot{\futurelet\@let@token\@onedot}

\newcommand{\figref}[1]{Fig\onedot~\ref{#1}}
\newcommand{\equref}[1]{Eq\onedot~\eqref{#1}}
\newcommand{\secref}[1]{Sec\onedot~\ref{#1}}
\newcommand{\tabref}[1]{Tab\onedot~\ref{#1}}

\def\eg{\emph{e.g}\onedot} 
\def\ie{\emph{i.e}\onedot}

\def\etal{\emph{et al}\onedot}

\title{High-Order Attention Models for Visual Question Answering}

\author{
   Idan Schwartz \\
   Department of Computer Science \\
   Technion\\
   \texttt{idansc@cs.technion.ac.il} \\
   \And
    Alexander G.~Schwing\\
    Department of Electrical and Computer Engineering\\
    University of Illinois at Urbana-Champaign\\
    \texttt{aschwing@illinois.edu} 
   \And
   Tamir Hazan\\
   Department of Industrial Engineering \& Management\\
   Technion\\
   \texttt{tamir.hazan@gmail.com}     
}
\begin{document}

\maketitle

% !TEX root = nips_2017.tex
\begin{abstract}
The quest for algorithms that enable cognitive abilities is an important part of machine learning. A common trait in many recently investigated cognitive-like tasks is that they  take into account different data modalities,  such as visual and textual input. In this paper we propose a novel and generally applicable form of attention mechanism that learns high-order correlations between various data modalities. We show that high-order correlations effectively direct the appropriate attention to the relevant elements in the different data modalities that are required to solve the joint task. We demonstrate the effectiveness of our high-order attention mechanism on the task of visual question answering (VQA), where we achieve state-of-the-art performance on the standard VQA dataset.

%Attention modules are a crucial component for reasoning tasks like machine translation, visual question answering (VQA), visual question generation, and image captioning. Beyond revealing interpretability, attention mechanisms have also demonstrated improved performance on the aforementioned tasks. However, careful development of attention mechanisms is still at its infancy. Here, we develop a general attention mechanism that can cope with any number of modalities. We demonstrate the effectiveness of our attention module on the VQA task where we demonstrate state-of-the-art performance. 

%Visual Question Answering (VQA) is a  tremendously challenging task as it combines computer vision, natural language processing and artificial intelligence. As such, it gives rise to numerous applications, for example for virtual assistance and computational education. Fortunately, recent advances in deep learning introduce successful implementations of VQA platforms using attention modules. These attention models are able to point out to semantic correlations between sentences and their images and perhaps more importantly, they improve the performance of VQA algorithms. In this paper we propose a technique that can both simplify the current attention models as well as cope with any number of modalities. We demonstrate the potential of our attention mechanism on VQA tasks, investigate its behavior, and evaluate its advantage on standard VQA benchmarks.
\end{abstract}

% !TEX root = nips_2017.tex
\section{Introduction}
The quest for algorithms  which enable cognitive abilities is an important part of machine learning and appears in many facets, \eg, in visual question answering tasks~\cite{das2016human}, image captioning~\cite{xu2015show}, visual question generation~\cite{mostafazadeh2016generating,JainZhangCVPR2017} and machine comprehension~\cite{hermann2015teaching}. A common trait in these recent cognitive-like tasks is that they  take into account different data modalities, for example, visual and textual data. 

To address  these tasks, recently, attention mechanisms have emerged as a powerful common theme, which provides not only some form of interpretability if applied to deep net models, but also often improves performance~\cite{hermann2015teaching}. The latter effect is attributed to more expressive yet concise forms of the various data modalities. Present day attention mechanisms, like for example~\cite{lu2016hierarchical,xu2015show}, are however often lacking in two main aspects. First, the systems generally extract abstract representations of data in an ad-hoc and entangled manner. Second, present day attention mechanisms are often geared towards a specific form of input and therefore hand-crafted for a particular task. %, \ie, without guiding principles. %to as is often ignored by simply combining representations of different data types before applying attention over the combined vector rather than both the individual as well as the combined representation separately. 

To address both issues, we propose a novel and generally applicable form of attention mechanism that learns high-order correlations between various data modalities. For example, second order correlations can model interactions between two data modalities, \eg, an image and a question, and more generally, $k-$th order correlations can model interactions between $k$ modalities. Learning these correlations effectively directs the appropriate attention to the relevant elements in the different data modalities that are required to solve the joint task. % `unary potentials,' based on individual data representations,  `pairwise potentials,' based on any combination of two data modalities, as well as `high order potentials,' based on three or more input data streams. %Second, we attempt to provide a gu.
%Our attention mechanism is based on a tensor approach to model correlations between a varying number of data modalities. %More specifically, we {\color{blue}xxx}.

\begin{figure}[t]
	\vspace{-0.7cm}
	\centering
	\includegraphics[width=1\linewidth]{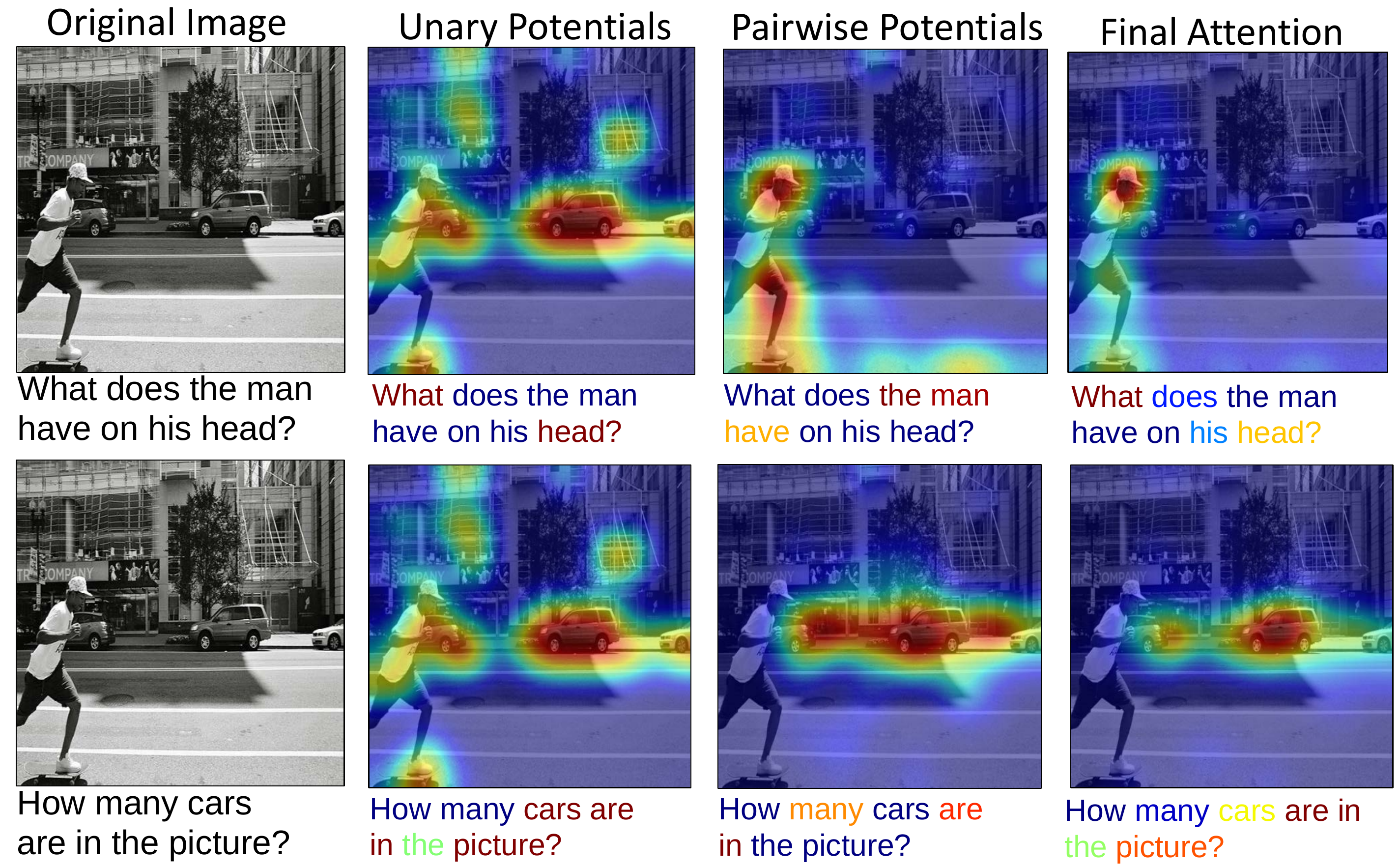}
	\vspace{-0.5cm}
	\caption{
		Results of our multi-modal attention for one image and two different questions (1$^\text{st}$ column). The unary image attention is identical by construction. 
		%The unary text attention focuses on the first word (2$^\text{nd}$ column). 
		The pairwise potentials differ for both questions and images since both modalities are taken into account (3$^\text{rd}$ column). The final attention is illustrated in the 4$^\text{th}$ column.
	}
	\label{fig:Teaser}
	\vspace{-0.2cm}
\end{figure}

We demonstrate the effectiveness of our novel attention mechanism on the task of visual question answering (VQA), where we achieve state-of-the-art performance on the VQA dataset~\cite{antol2015vqa}. Some of our results are visualized in \figref{fig:Teaser}, where we show how the visual attention correlates with the textual attention.

We begin by reviewing the related work. We subsequently provide details of our proposed technique, focusing on the high-order nature of our attention models. We then conclude by presenting the application of our high-order attention mechanism to VQA and compare it to the state-of-the-art. 

% !TEX root = nips_2017.tex
\section{Related work}
Attention mechanisms have been investigated for both image and textual data. In the following we review mechanisms for both. % in the following.

\noindent\textbf{Image attention mechanisms:} Over the past few years, single image embeddings extracted from a deep net (\eg,~\cite{malinowski2015ask,ma2015learning}) have been extended to a variety of image attention modules, when considering VQA. For example, a textual long short term memory net (LSTM) may be augmented with a spatial attention~\cite{zhu2016visual7w}. Similarly, Andreas \etal~\cite{andreas2016learning} employ a language parser together with a series of neural net modules, one of which attends to regions in an image. The language parser suggests which neural net module to use. Stacking of attention units was also investigated by Yang \etal~\cite{yang2016stacked}. Their stacked attention network predicts the answer successively. 
Dynamic memory network modules which capture contextual information from neighboring image regions has been considered by Xiong \etal~\cite{xiong2016dynamic}.
Shih \etal~\cite{shih2016look}  use object proposals and and rank regions according to relevance. The multi-hop attention scheme of Xu \etal~\cite{xu2016ask} was proposed to extract fine-grained details. A joint attention mechanism was discussed by Lu \etal~\cite{lu2016hierarchical} and Fukui \etal~\cite{fukui2016multimodal} suggest an efficient outer product mechanism to combine visual representation and text representation before applying attention over the combined representation. 
Additionally, they suggested the use of glimpses. Very recently, Kazemi \etal~\cite{kazemi2017show} showed a similar approach using concatenation instead of outer product.
Importantly, all of these approaches model attention as a single network.  The fact that multiple modalities are involved is often not considered explicitly which contrasts the aforementioned approaches from the technique we present.

Very recently Kim \etal~\cite{kim2017structured}  presented a technique that also interprets attention as a multi-variate probabilistic model, to incorporate structural dependencies into the deep net. Other recent techniques are work by Nam \etal~\cite{nam2016dual} on dual attention mechanisms and work by Kim \etal~\cite{kim2016hadamard} on bilinear models. In contrast to the latter two models our approach is easy to extend to any number of data modalities.

%{\color{red} again I think this part is not crucial}	
%{\color{green}
\noindent\textbf{Textual attention mechanisms:} We also want to provide a brief review of textual attention. 
To address some of the challenges, \eg, long sentences, faced by translation models, Hermann \etal~\cite{hermann2015teaching} proposed RNNSearch. To address the challenges which arise by fixing the latent dimension of neural nets processing text data, Bahdanau \etal\cite{bahdanau2014neural} first encode a document and a query via a bidirectional LSTM which are then used to compute attentions. This mechanism was later refined in~\cite{rocktaschel2015reasoning} where a word based technique reasons about sentence representations. Joint attention between two CNN hierarchies is discussed by Yin \etal~\cite{yin2015abcnn}.

Among all those attention mechanisms,  relevant to our approach is work by Lu \etal~\cite{lu2016hierarchical} and the approach presented by Xu \etal~\cite{xu2016ask}. Both discuss attention mechanisms which operate jointly over two modalities.  Xu \etal~\cite{xu2016ask} use pairwise interactions in the form of a similarity matrix, but ignore the attentions on individual data modalities. %, which have substantial benefit. Moreover, 
Lu \etal~\cite{lu2016hierarchical} suggest an alternating model, that directly combines the features of the modalities before attending. % and repeats the process similarly to EM algorithms. 
Additionally, they suggested a parallel model which uses a similarity matrix to map features for one modality to the other.  It is hard to extend this approach to more than two modalities. In contrast, our model  develops a probabilistic model, based on high order potentials and performs mean-field inference to obtain marginal probabilities. This permits trivial extension of the model to  any number of modalities.

Additionally, Jabri \etal~\cite{jabri2016revisiting} propose a model where answers are also used as inputs. Their approach questions the need of attention mechanisms and develops an alternative solution based on binary classification. In contrast, our approach captures high-order attention correlations, which we found to improve performance significantly.

Overall, while there is early work that propose a combination of language and image attention for VQA, \eg,~\cite{lu2016hierarchical,xu2016ask,kim2016multimodal}, attention mechanism with several potentials haven't been discussed in detail yet. In the following we present our approach for joint attention over any number of modalities.

% !TEX root = nips_2017.tex
\section{Higher order attention models}

Attention modules are a crucial component for present day decision making systems. Particularly when taking into account more and more data of different modalities, attention mechanisms are able to provide  insights into the inner workings of the oftentimes abstract and automatically extracted representations of our systems.

An example of such a system that captured a lot of research efforts in recent years is Visual Question Answering (VQA). Considering VQA as an example, we immediately note its dependence on two or even three different data modalities, the visual input $V$, the question $Q$ and the answer $A$, which get processed simultaneously. More formally, we let
\bes
V\in\mathbb{R}^{n_{v}\times d}, \quad
Q\in\mathbb{R}^{n_{q}\times d},\quad
A\in\mathbb{R}^{n_{a}\times d}
\ees  
denote a representation for the visual input, the question and the answer respectively. Hereby, $n_v$, $n_q$ and $n_a$ are the number of pixels, the number of words in the question, and the number of possible answers. We use $d$ to denote the dimensionality of the data. For simplicity of the exposition we assume $d$ to be identical across all data modalities.

%We use VQA  as an example; here is our general approach; what is the data

\begin{figure}[t]
\centering
\hspace{-4.5cm}
	\includegraphics[width=0.7\linewidth]{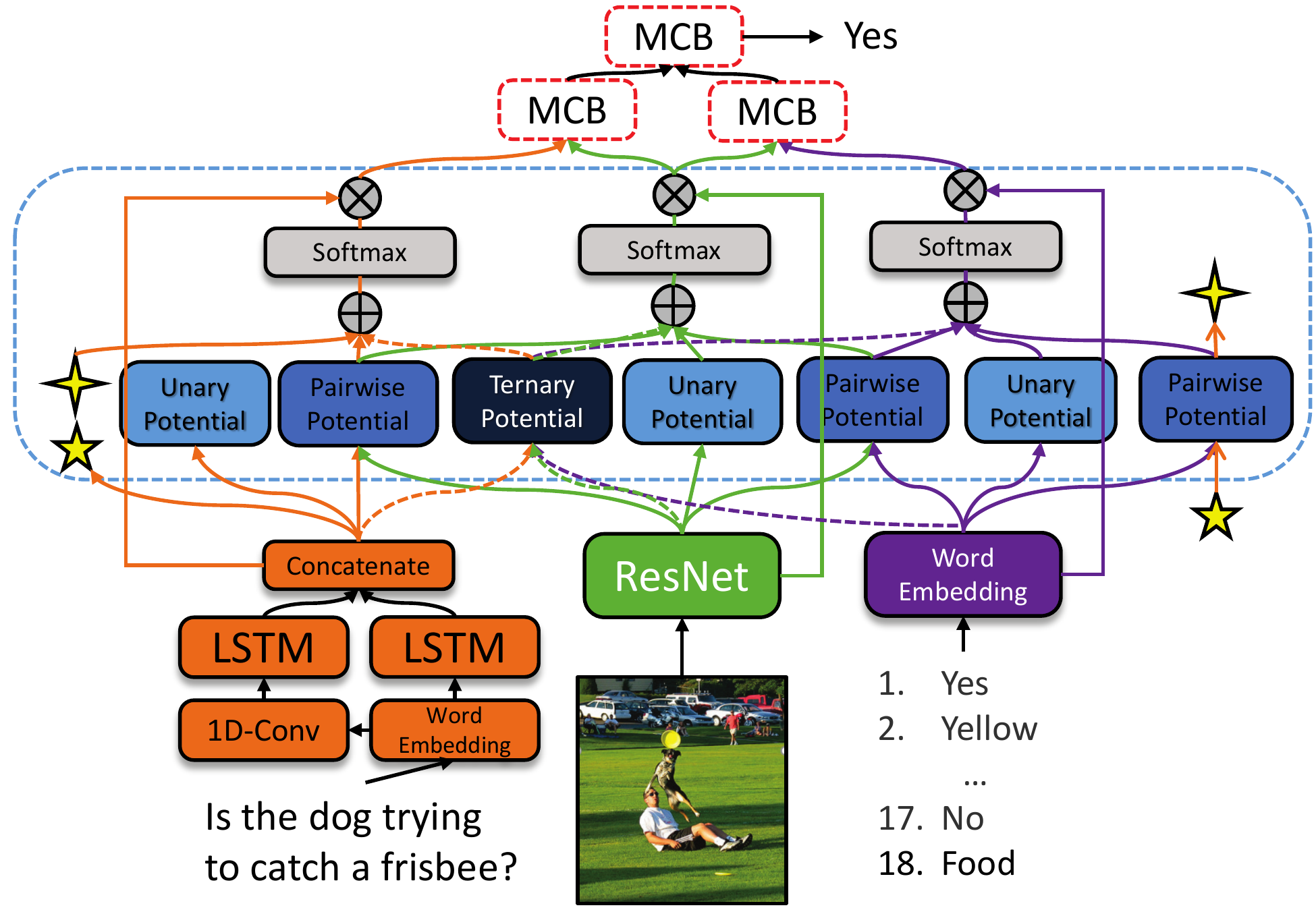}
	\put(-15,45){$\left.\begin{array}{c}~\\~\\~\\~\\~\\~\\~\end{array}\right\}$ Data embedding (\secref{sec:DataEmbed})}
	\put(-15,120){$\left.\begin{array}{c}~\\~\\~\\~\\~\\~\\\end{array}\right\}$ Attention (\secref{sec:Att})}
	\put(-15,170){$\left.\begin{array}{c}~\\~\\\end{array}\right\}$ Decision (\secref{sec:DecMaking})}
\caption{Our state-of-the-art VQA system}
\label{fig:VQASys}
\end{figure}

%Three parts: data embedding, attention, decision making. Subsequently we discuss each of those parts more carefully.
Due to this dependence on multiple data modalities, present day decision making systems can be decomposed into three major parts: (i) the data embedding; (ii) attention mechanisms; and (iii) the decision making. For a state-of-the-art VQA system such as the one we developed here, those three parts are immediately apparent when considering the high-level system architecture outlined in \figref{fig:VQASys}.

\subsection{Data embedding}
\label{sec:DataEmbed}

Attention modules deliver to the decision making component a succinct representation of the relevant data modalities. As such, their performance depends on how we represent the data modalities themselves. Oftentimes, an attention module tends to use expressive yet concise data embedding algorithms to better capture their correlations and consequently to improve the decision making performance. For example, data embeddings based on convolutional deep nets which constitute the state-of-the-art in many visual recognition and scene understanding tasks. Language embeddings heavily rely on LSTM which are able to capture context  in sequential data, such as words, phrases and sentences. We give a detailed account to our data embedding architectures for VQA in \secref{sec:Embed}.

\subsection{Attention}
\label{sec:Att}
As apparent from the aforementioned description, attention is the crucial component connecting data embeddings with decision making modules.

Subsequently we denote attention over the $n_q$ words in the question via $P_Q(i_q)$, where $i_q\in\{1, \ldots, n_q\}$ is the word index. Similarly, attention over the image is referred to via $P_V(i_v)$, where $i_v\in\{1, \ldots, n_v\}$, and attention over the  possible answers are denoted $P_A(i_a)$, where $i_a\in\{1, \ldots, n_a\}$.

We consider the attention mechanism as a probability model, with each attention mechanism computing ``potentials.'' First,  unary potentials $\theta_V$, $\theta_Q$, $\theta_A$ denote the importance of each feature (\eg, question word representations, multiple choice answers representations, and  image patch features) for the VQA task. Second,  pairwise potentials, $\theta_{V,Q}, \theta_{V,A}, \theta_{Q,A}$ express correlations between two modalities. Last,  third-order potential, $\theta_{V,Q,A} $ captures dependencies between the three modalities. %Three modalities is evaluated over the task of Multiple Choice Answering, where in addition to the question and the image, a set of possible answers is added as a third modality. Therefore, $\theta_{V,Q,A}$ expressing the correlation between a possible answer, a question's word, and image pixel.  

%For simplicity, we will start by describing a two modalities solution, where in \secref{HighOrder} we will explain how to apply a third modality. 
To obtain marginal probabilities $P_Q$, $P_V$ and $P_A$ from potentials, our model performs mean-field inference. We combine the unary potential, the marginalized pairwise potential and the marginalized third order potential linearly including a bias term:
%\beas
%P_V(i_v) &=& \operatorname{smax} \left(\alpha_1\theta_V(i_v) + \alpha_2\theta_{V,Q}(i_v) + \alpha_3 \right), \\ 
%P_Q(i_q) &=& \operatorname{smax}\left(\beta_1\theta_Q(i_q) + \beta_2\theta_{V,Q}(i_q) + \beta_3 \right).
%\eeas
\bea
P_V(i_v) &=& \operatorname{smax}\!\left(\! \alpha_1\theta_V(i_v) \!+\! \alpha_2\theta_{V,Q}(i_v) \!+\! \alpha_3\theta_{A,V}(i_v) \!+\! \alpha_4\theta_{V,Q,A}(i_v) + \alpha_5 \!\right)\!,\nonumber\\
P_Q(i_q) &=& \operatorname{smax}\!\left(\! \beta_1\theta_Q(i_q) \!+\! \beta_2\theta_{V,Q}(i_q) \!+\! \beta_3\theta_{A,Q}(i_q) \!+\! \beta_4\theta_{V,Q,A}(i_q) + \beta_5 \!\right)\!, \label{eq:AttentionProbs}\\
P_A(i_a) &=& \operatorname{smax}\!\left(\! \gamma_1\theta_A(i_a) \!+\! \gamma_2\theta_{A,V}(i_a) \!+\! \gamma_3\theta_{A,Q}(i_a) \!+\! \gamma_4\theta_{V,Q.A}(i_a) + \gamma_5 \!\right)\!. \nonumber
\eea
Hereby $\alpha_i$, $\beta_i$, and $\gamma_i$ are learnable parameters and $\operatorname{smax}(\cdot)$ refers to the soft-max operation over $i_v\in\{1, \ldots, n_v\}$, $i_q\in\{1, \ldots, n_q\}$ and $i_a\in\{1, \ldots, n_a\}$ respectively. The soft-max converts the combined potentials to probability distributions,  which corresponds to a single mean-field iteration.
Such a linear combination of potentials provides extra flexibility for the model, since it can learn the reliability of the potential from the data.  For instance,  we observe that question attention relies more on the unary question potential and on pairwise question and answer potentials. In contrast, the image attention relies more on the pairwise question and image potential.  %the more important potential.  %{\color{blue}We found this extra flexibility to yield ...} {\color{red} ( shouldn't this go to ablation study?) interesting insights, considering the coefficients, 
  %suggesting that the model doesn't need the picture to understand which are the more important words} Afterwards we use a softmax, which converts the potentials into Gibbs distributions. More formally, we obtain the attended features {\color{blue}this is what $a_v$ was used for earlier. is this still the case? also make sure that the subscripts are identical, \ie, either lower-case or upper-case}

Given the aforementioned probabilities $P_V$, $P_Q$, and $P_A$, the attended image, question and answer vectors are denoted by $a_V\in\mathbb{R}^{d}$, $a_Q\in\mathbb{R}^{d} $ and $a_A\in\mathbb{R}^{d}$. The attended modalities are calculated as the weighted sum of the image features $V = [v_1, \ldots, v_{n_v}]^T\in\mathbb{R}^{n_v\times d}$, the question features $Q = [q_1, \ldots, q_{n_q}]^T\in\mathbb{R}^{n_q\times d}$, and the answer features $A = [a_1, \ldots, a_{n_a}]^T\in\mathbb{R}^{n_a\times d}$, \ie,
\bes
a_V = \sum_{i_v = 1}^{n_v} P_V(i_v)v_{i_v},  \quad  a_Q = \sum_{i_q=1}^{n_q} P_Q(i_q) q_{i_q}, \quad \text{and} \quad a_V = \sum_{i_a=1}^{n_a} P_A(i_a) a_{i_a}.
\ees
The attended modalities, which effectively focus on the  data relevant for the  task, are passed to a classifier for decision making, \eg, the ones discussed in \secref{sec:DecMaking}. %\color{blue} either short description here or later in a separate section. 
%For an illustration of this pairwise mechanism please refer to \figref{fig:pw_fig}. {\color{blue}which figure is this?}
%
In the following we now describe the attention mechanisms for unary, pairwise and  ternary potentials in more detail. 

\vspace{-0.2cm}
\subsubsection{Unary potentials}
\label{sec:UnaryAttention}
\vspace{-0.2cm}

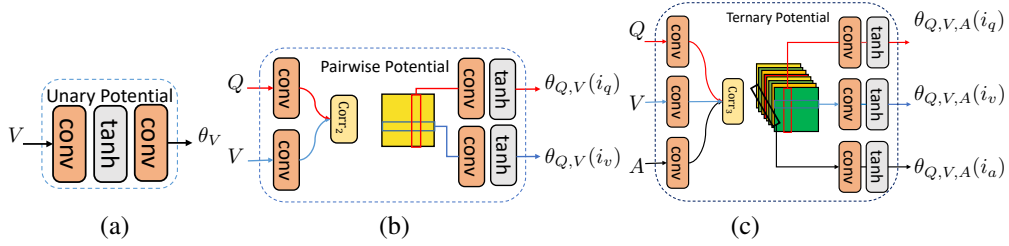
\begin{figure}[t]
	\vspace{0.6cm}
	\centering
\begin{tabular}{ccc}
\hspace{-0.8cm}
\scalebox{0.8}{\includegraphics[width=3cm]{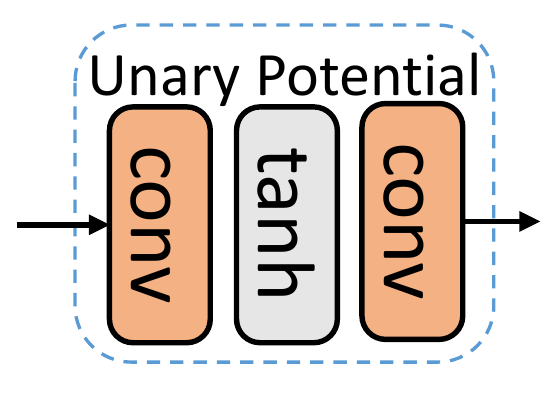}
	\put(-90,30){$V$}
	\put(-0,30){$\theta_V$}
	
    } 	\hspace{0.1cm}
    &
    \scalebox{0.8}{\includegraphics[width=5cm]{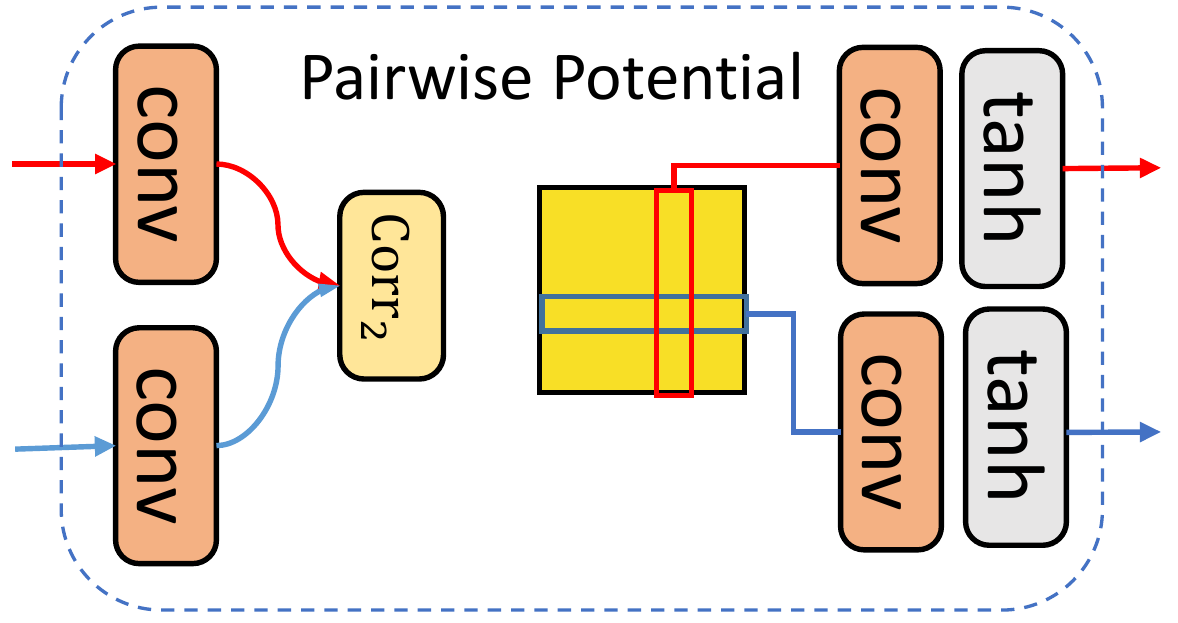}
    	\put(-150,55){$Q$}
    	\put(-150,20){$V$}
    	\put(0,55){$\theta_{Q,V}(i_q)$}
    	\put(0,20){$\theta_{Q,V}(i_v)$}}
    &
	\hspace{0.8cm}
    \scalebox{0.8}{\includegraphics[width=4.5cm]{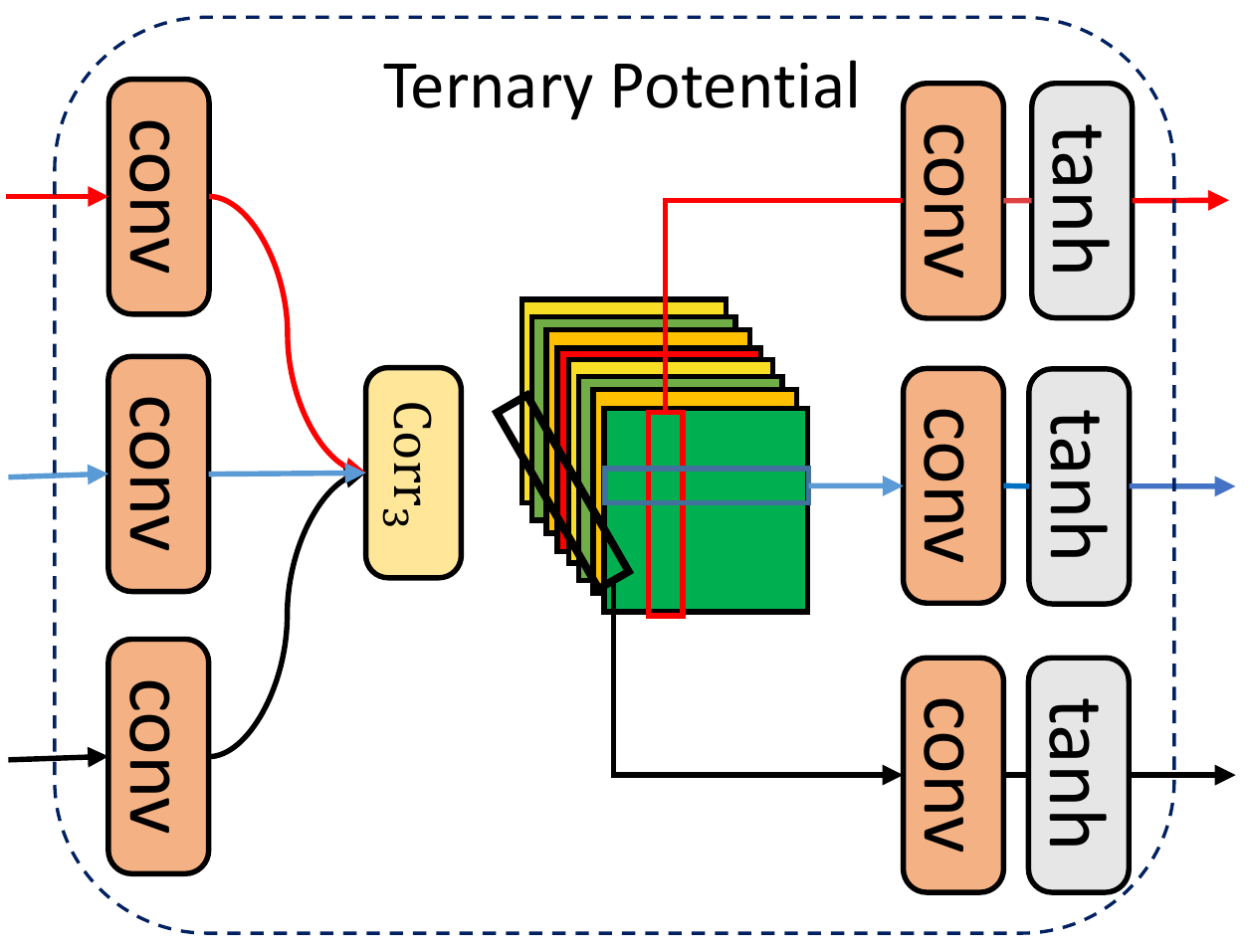}
    	\put(-135,80){$Q$}
    	\put(-135,45){$V$}
    	\put(-135,15){$A$}
    	\put(0,85){$\theta_{Q,V,A}(i_q)$}
    	\put(0,50){$\theta_{Q,V,A}(i_v)$}
    	\put(0,15){$\theta_{Q,V,A}(i_a)$}}
    \\
   \hspace{-0.8cm} (a)&(b)&(c)
\end{tabular}
	\vspace{-0.2cm} 
	\caption{\small  Illustration of our $k-$order attention. (a) unary attention module (e.g., visual). (b) pairwise attention module (e.g., visual and question) marginalized over its two data modalities. (c) ternary attention module (e.g., visual, question and answer) marginalized over its three data modalities..
		}
	\label{fig:un_fig}
	\vspace{-0.5cm}
\end{figure}

%Subsequently we describe our unary attention module more formally. 
We illustrate the unary attention schematically in \figref{fig:un_fig}~(a). The input to the unary attention module is a data representation, \ie, either the visual representation $V$, the question representation $Q$, or the answer representation $A$.
Using those representations, we obtain the `unary potentials' $\theta_V$, $\theta_Q$ and $\theta_A$ using a convolution operation with kernel size $1\times 1$ over the data representation as an additional embedding step, followed by a non-linearity ($\tanh$ in our case), followed by another convolution operation with kernel size $1\times 1$ to reduce embedding dimensionality. Since convolutions with kernel size $1\times 1$ are identical to matrix multiplies we formally obtain the unary potentials via
\bes
\theta_V(i_v) =  \tanh(V W_{v_2} ) W_{v_1}, \quad \theta_Q(i_q) =  \tanh(Q W_{q_2} ) W_{q_1}, \quad 
\theta_A(i_a) =  \tanh(A W_{a_2} ) W_{a_1}.
\ees
%We omitted the bias terms for notational simplicity.
% {\color{blue} are those matrix multiplies? I guess we should describe and include a convolution operator?}{\color{red} not sure, since a 1x1 conv is basically like multiplying a matrix am I wrong? }
where $W_{v_1}, W_{q_1} , W_{a_1}\in\mathbb{R}^{d\times 1}$, and $W_{v_2}, W_{q_2}, W_{a_2} \in \mathbb{R}^{d\times d}$ are trainable parameters.%, $d$ is embedding dimension, $ V\in\mathbb{R}^{n_v \times d}$, $Q\in \mathbb{R}^{n_q \times d} $ and $A\in\mathbb{R}^{n_a\times d}$ are the image, question and answer representation. %, setting $d_2 < d_1$  allowing regularization %{\color{blue} some more details and how it relates to the figure} {\color{red} I'm still not sure how to justify that, it was suggested by Tamir in the past so maybe there is an actual paper that we can cite }

%{\color{blue} how about bias terms?}
%{color{red} Added, but maybe we should just write we omit bias terms to make things more readable?}

\vspace{-0.2cm}
\subsubsection{Pairwise potentials}
\label{sec:PairAttention}
\vspace{-0.2cm}

Besides the mentioned mechanisms to generate unary potentials, we specifically aim at taking advantage of pairwise attention modules, which are able to capture the correlation between the representation of different modalities.  Our approach is illustrated in \figref{fig:un_fig}~(b). We use a similarity matrix between image and question modalities $C_2 = Q W_q (VW_v)^\top $. Alternatively, the $(i,j)$-th entry is the correlation (inner-product) of the $i$-th column of $QW_q $  and the $j$-th column of $VW_v$:  
\bes
(C_2)_{i,j} = \operatorname{corr_2}((QW_q)_{:,i}, (VW_v)_{:,j}), \quad\quad \operatorname{corr_2}(q,v) = \sum_{l=1}^d q_l v_l.
\ees
%$ \operatorname{corr_2}(\cdot) $  is computed using dot product between vectors. %{\color{blue} dimensions x,y should be specified}. 
where $W_q,W_v\in\mathbb{R}^{d\times d}$ are trainable parameters. %Intuitively, the matrix entry $(C_2)_{i,j}$ represents the correlation between the vectors $q_i$ and $v_j$ corresponding to the  $i$-th word in the question and the $j$-th pixel representation in the image. 
We consider $(C_2)_{i,j}$ as a pairwise potential that represents the correlation of the $i$-th word in a question and the $j$-th patch in an image. Therefore, to retrieve the attention for a specific word, we convolve the matrix along the visual dimension using a  $1\times 1$ dimensional kernel.  Specifically, 
\bes
\theta_{V,Q}(i_q) = \tanh\left(\sum_{ i_v=1}^{n_v}  w_{i_v} (C_2)_{i_v,i_q}\right), \quad\text{and}\quad \theta_{V,Q}(i_v) = \tanh\left( \sum_{ i_q = 1}^{n_q} w_{i_q}  (C_2)_{i_v,i_q} \right).
\ees
%{\color{blue}the blue colored indices need to be adjusted}
Similarly, we obtain $\theta_{A,V}$ and $\theta_{A,Q}$, which we omit due to space limitations. These potentials are used to compute the attention probabilities as defined in \equref{eq:AttentionProbs}.

%{\color{blue} ref to \figref{fig:pw_fig} missing}
%{\color{blue} the figure contains operations (conv,tanh) at the output that aren't described here}
%{\color{red} Conv is multiplying by matrix, since it's 1x1 filter}
%{\color{blue}how about bias terms}

\subsubsection{Ternary Potentials}

%While pairwise attention is beneficial for two modalities it is unclear how correlation based techniques can be extended to any number of data modalities. %An alternative task to visual question answering is multiple choice visual question answering. In this setup, a set of possible answers is added to the given question and image. Following this setup, the possible answers are considered as the $3^\text{rd}$ modality. We denote 
%\bes
%A\in\mathbb{R}^n_a\times d
%\ees 
%where $n_a$, is the number of possible answers. the attention over the $n_a$ possible answers via $P_A(i_a)$, where $i_a\in\{1, \ldots, n_a\}$ is the answer index. 
To capture the dependencies between all three modalities, we consider their high-order correlations.
%To capture the interaction of three vectors  $ q_i,v_j,a_k\in\mathcal{R}^{d} $ we define the operator 
\beas
(C_3)_{i,j,k} = \operatorname{corr_3}((QW_q)_{:,i}, (VW_v)_{:,j}, (AW_a)_{:,k}), \quad\quad \operatorname{corr_3}(q,v,a) = \sum_{l=1}^d q_l v_l  a_l.
\eeas

%Based on this operator we can define a high-order matching between tensors. More formally, given the question representation $Q\in\mathbb{R}^{n_q\times d} $, the image representation $V\in\mathbb{R}^{n_v\times d}$ as well as the answer representation $A\in\mathbb{R}^{n_a\times d} $  the dependency is computed according to 
%\beas
%C_3 = \operatorname{Corr_3}\left(W_q Q, W_vV, W_aA\right) \in \mathbb{R}^{n_q\times n_v\times n_a}
%\eeas
%where the  $ i,j,k $-th entry of $ C_3 $ captures the joint interaction of vectors $q_i,v_j,a_k$ specified by $ \operatorname{corr_3}(q_i,v_j,a_k) = \sum_{n=1,\ldots, d} q_{i,n}\cdot v_{j,n}\cdot a_{k,n} $.  For notational simplicity we omitted the bias terms. % for notational simplicity. %{\color{blue} dimensions x,y should be specified}. 
%
%To conclude, each cell in $C_3$ is defined accordingly: 
%
%$$
%(C_3)_{i,j,k} = \operatorname{corr_3}((W_qQ)_{:,i}, (W_vV)_{:,j}, (W_aA)_{:,k}),
%$$
%where the  $ i,j,k $-th entry of $ C_3 $, \ie, $(C_3)_{i,j,k}$, captures the joint interaction of vectors $q_i,v_j,a_k$.

Where $W_q,W_v,W_a\in\mathbb{R}^{d\times d}$ are trainable parameters. Similarly to the pairwise potentials, we use the $C_3$ tensor to obtain correlated attention for each modality:
\bes
\theta_{V,Q,A}(i_q) \!=\! \tanh\!\left(\sum_{ i_v = 1}^{n_v} \sum_{i_a=1}^{n_a}  w_{i_v,i_a}  (C_3)_{i_q,i_v,i_a} \!\right)\!, \;
 \theta_{V,Q,A}(i_v) \!=\! \tanh\!\left(\sum_{ i_q = 1}^{n_q} \sum_{i_a=1}^{n_a} w_{i_q,i_a}  (C_3)_{i_q,i_v,i_a}\!\right)\!
\ees
\bes
\text{and} \quad\theta_{V,Q,A}(i_a) = \tanh\left( \sum_{ i_v=1}^{n_v} \sum_{i_q=1}^{n_q} w_{i_q,i_a}  (C_3)_{i_q,i_v,i_a} \right).
\ees

These potentials are used to compute the attention probabilities as defined in \equref{eq:AttentionProbs}.

\subsection{Decision making}
\label{sec:DecMaking}

The decision making component receives as  input the attended modalities and predicts the desired output. Each attended modality is a vector that consists of the relevant data for making the decision. While the decision making component can consider the modalities independently, the nature of the task usually requires to take into account correlations between the attended modalities. The correlation of a set of attended modalities are represented by the outer product of their respective vectors, \eg, the correlation of two attended modalities  is represented by a matrix and the correlation of $k$-attended modalities is represented by a $k$-dimensional tensor. 

\begin{figure}[t]
	\vspace{-0.6cm}
	\centering
	\begin{tabular}{ccc}
		\scalebox{0.8}{\includegraphics[width=4cm]{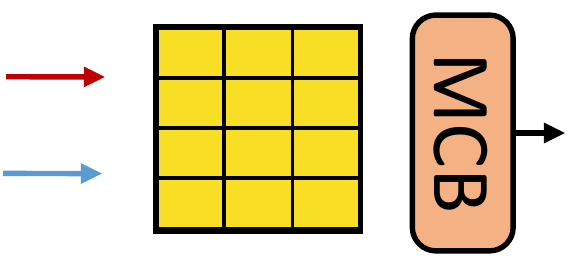}
			\put(-125,40){$a_Q$}
			\put(-125,17){$a_V$}					
			
		} 
		&&
		\scalebox{0.8}{\includegraphics[width=5cm]{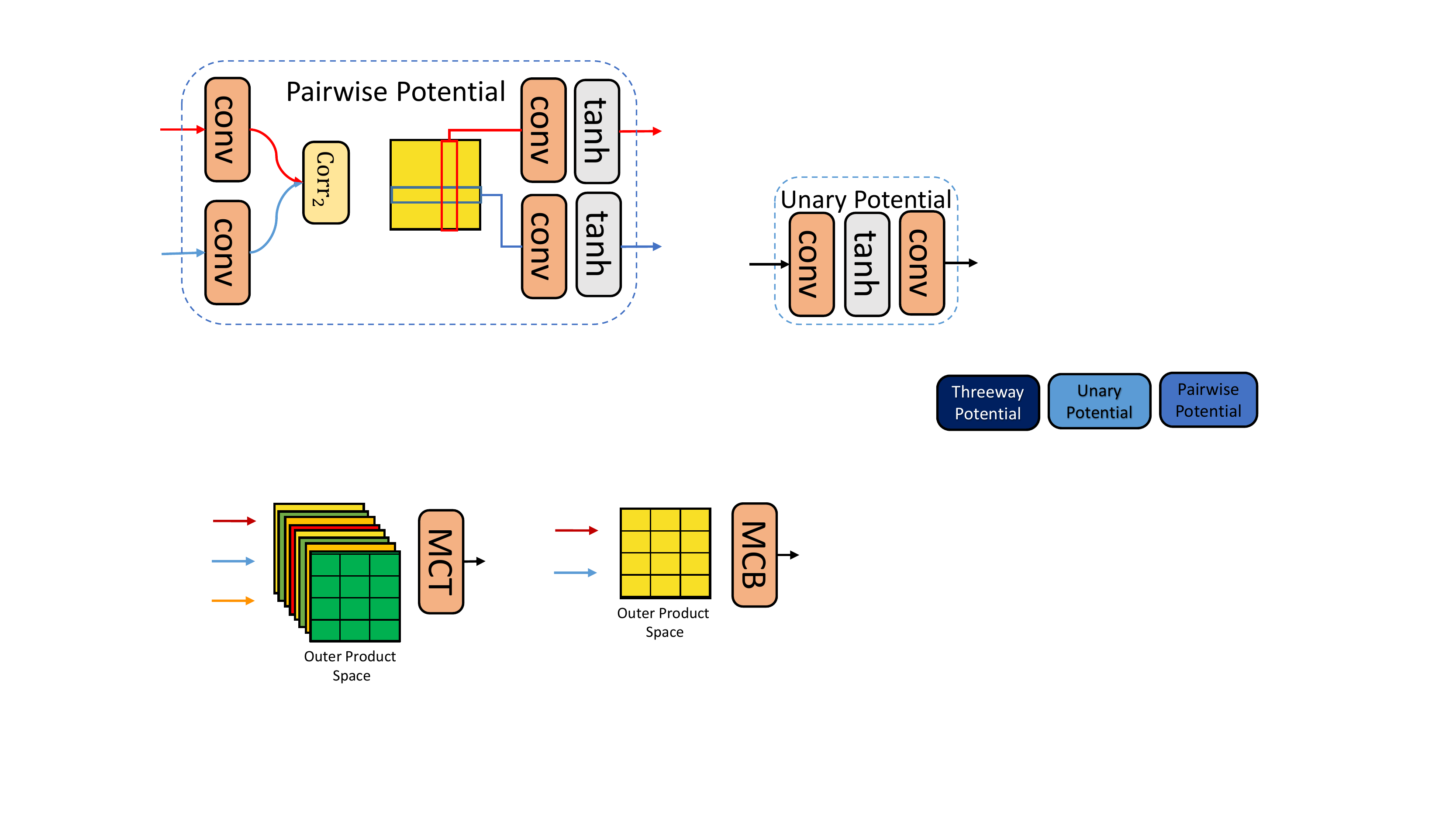}
			\put(-155,65){$a_Q$}
			\put(-155,42){$a_V$}
			\put(-155,22){$a_A$}	
		}
		\\
		(a)&&(b)
	\end{tabular}
	\vspace{-0.2cm}
	\caption{\small 
		{Illustration of correlation units used for decision making. (a) MCB unit approximately sample from outer product space of two attention vectors, (b) MCT unit approximately sample from outer product space of three attention vectors.}
	}
	\label{fig:mcb_tcb}
	\vspace{-0.5cm}
\end{figure}
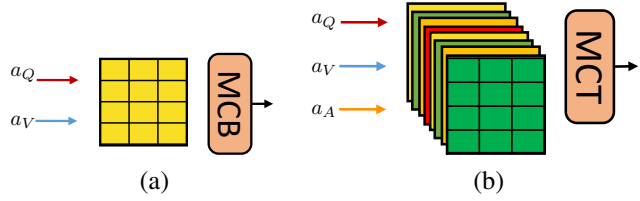

Ideally, the attended modalities and their high-order correlation tensors are fed into a  deep net which produces the final decision. The number of parameters in such a network grows exponentially in the number of modalities, as seen in \figref{fig:mcb_tcb}. To overcome this computational bottleneck, we follow the tensor sketch algorithm of Pham and Pagh~\cite{pham2013fast}, which was recently applied to attention models by Fukui \etal~\cite{fukui2016multimodal} via  Multimodal Compact Bilinear Pooling (MCB) in the pairwise setting or Multimodal Compact Trilinear Pooling (MCT),  an extension of MCB that pools data from three modalities. The tensor sketch algorithm enables us to reduce the dimension of any rank-one tensor while referring to it implicitly. It relies on the count sketch technique~\cite{charikar2002finding} that randomly embeds an attended vector $a \in \mathbb{R}^{d_1}$ into another Euclidean space $\Psi(a) \in \mathbb{R}^{d_2}$. The tensor sketch algorithm then projects the rank-one tensor $\otimes_{i=1}^k a_i$ which consists of attention correlations of order $k$ using the convolution $\Psi(\otimes_{i=1}^k a_i ) = \ast_{i=1}^k  \Psi(a_i )$. For example, for two attention modalities, the correlation matrix $a_1 a_2^\top = a_1 \otimes a_2$ is randomly projected to $\mathbb{R}^{d_2}$ by the convolution $\Psi(a_1 \otimes a_2) = \Psi(a_1) \ast \Psi(a_2)$. The attended modalities $\Psi(a_i)$ and their high-order correlations $\Psi(\otimes_{i=1}^k a_i )$ are fed into a fully connected neural net to complete  decision making.

%\input{intro}
%\input{rel}
% !TEX root = nips_2017.tex

\section{Visual question answering}

In the following we evaluate our approach qualitatively and quantitatively. Before doing so we describe the data embeddings.

\subsection{Data embedding}
\label{sec:Embed}
The attention module requires the question representation $Q\in\mathbb{R}^{n_q\times d}$,  the image representation $V\in\mathbb{R}^{n_v\times d}$, and the answer representation $A\in\mathbb{R}^{n_a\times d}$, which are computed as follows. %We next detail how to compute those.

\noindent\textbf{Image embedding:} 
To embed the image, we use pre-trained convolutional deep nets (\ie, VGG-19, ResNet). We extract the last layer before the fully connected units. Its dimension in the VGG net case is $512\times14\times14$ and the dimension in the ResNet case is $2048\times14\times14$. Hence we obtain $n_v = 196$ and we embed both the 196 VGG-19 or ResNet features into a $d=512$ dimensional space to obtain the image representation $V$. 

\noindent\textbf{Question embedding:} 
To obtain a question representation, $Q\in\mathcal{R}^{n_q\times d}$, we first map a 1-hot encoding of each word in the question into a $d$-dimensional embedding space using a linear transformation plus corresponding bias terms. To obtain a richer representation that accounts for neighboring words, we use a 1-dimensional temporal convolution with filter of size 3. While a combination of multiple sized filters is suggested in the literature~\cite{lu2016hierarchical}, we didn't find any  benefit from using such an approach.  Subsequently, to capture long-term dependencies, we used a Long Short Term Memory (LSTM) layer. To reduce overfitting caused by the LSTM units, we used two LSTM layers with $d/2$ hidden dimension, one uses as input the word embedding representation, and the other one operates on the 1D conv layer output. Their output is then concatenated to obtain $Q$. % as illustrated in \figref{fig:VQAs}. 
We also note that $n_q$ is a constant hyperparameter, \ie, questions with more than $n_q$ words are cut, while questions with less words are zero-padded.

\noindent\textbf{Answer embedding:} 
To embed the possible answers we use a regular word embedding. The vocabulary is specified by taking only the most frequent answers in the training set. Answers that are not included in the top answers are embedded to the same vector. Answers containing multiple words are embedded as n-grams to a single vector. We assume there is no real dependency between the answers, therefore there is no need of using additional 1D conv, or LSTM layers.

\begin{table}[t]
\vspace{-0.5cm}
	\setlength{\tabcolsep}{5pt}
	\centering
		\caption{Comparison of results on the Multiple-Choice VQA dataset for a variety of methods. We observe the combination of all three unary, pairwise and ternary potentials to yield the best result.}
	\begin{tabular}{lcccccc}
		\toprule
		\multicolumn{1}{c}{}   & \multicolumn{4}{c}{test-dev}  & \multicolumn{1}{c}{test-std}   & \\
		\cmidrule(r){2-5}
		\cmidrule(r){6-7}
		{Method} & Y/N  & Num & Other & All & All \\ 	  \midrule
		HieCoAtt (VGG)~\cite{lu2016hierarchical}
		& {79.7} & {40.1} & {57.9} & {64.9} & {-}  \\		
		HieCoAtt (ResNet)~\cite{lu2016hierarchical}
		& {79.7} & {40.0} & {59.8} & {65.8} & {66.1}  \\
		{RAU (ResNet)~\cite{noh2016training}}		
		& {81.9} & {41.1} & {61.5} & {67.7} & {67.3} \\
		{MCB (ResNet)~\cite{fukui2016multimodal}}
		& {-} & {-} & {-} & {68.6} & {-} \\
		{DAN\ (VGG)~\cite{nam2016dual}}		
		& {-} & {-} & {-} & {67.0} & {-} \\
		{DAN\ (ResNet)~\cite{nam2016dual}}
		& {-} & {-} & {-} & {69.1} & {69.0} \\
		{MLB\ (ResNet)~\cite{kim2016hadamard}}
		& {{-}} & {-} & {{-}} & {{-}} & {68.9} \\
		\midrule
		{2-Modalities: Unary+Pairwis\ (ResNet)}& {80.9} & {36.0} & {61.6} & {66.7} & {-}\\ 
		{3-Modalities: Unary+Pairwise\ (ResNet)}& \bf{{82.0}} & {42.7} & {63.3} & {68.7} & {68.7} \\
		{3-Modalities: Unary + Pairwise + Ternary\ (VGG)} & {81.2} & {42.7} & {62.3} & {67.9} &  {-}  \\
		{3-Modalities: Unary + Pairwise + Ternary\ (ResNet)} & {81.6} & \bf{{43.3}} & \bf{{64.8}} & \bf{{69.4}} &  \bf{{69.3}}  \\
		
		\bottomrule
	\end{tabular}

	\label{tab:compareattention}
	\vspace{-0.4cm}
\end{table}

\subsection{Decision making}
\label{sec:DecisionExp}
For our VQA example we investigate two techniques to combine vectors from three modalities. First, the attended feature representation for each modality, \ie, $a_V$, $a_A$ and $a_Q$,  are combined using an MCT unit. Each feature element is  of the form $ ((a_V)_i\cdot(a_Q)_j\cdot(a_A)_k) $. While this first solution  is most general, in some cases like VQA, our experiments show that it is better to use our second approach, a 2-layer MCB unit combination. This permits  greater expressiveness as we employ features of the form $((a_V)_i\cdot(a_Q)_j\cdot(a_Q)_k\cdot(a_A)_t)$ therefore also allowing image features to interact with themselves. Note that in terms of parameters both approaches are identical as neither MCB nor MCT are  parametric modules.

Beyond MCB, we tested several other techniques that were suggested in the literature, including element-wise multiplication, element-wise addition and concatenation~\cite{kim2016hadamard, lu2016hierarchical, kazemi2017show}, optionally followed by another hidden fully connected layer. The tensor sketching units consistently performed best. % result.

% !TEX root = nips_2017.tex

%\vspace{0cm}
\subsection{Results}
%\vspace{1cm}

\begin{figure*}[t]
	\centering
	\setlength\tabcolsep{0cm}
	\begin{tabular}{ccccccc}
		\raisebox{0.1cm}{\includegraphics[clip=true,width=1.8cm,height=1.8cm]{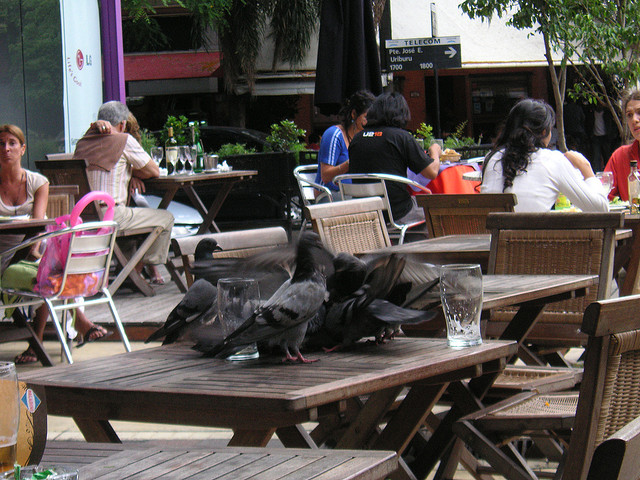}}& 
		\includegraphics[clip=true,width=0.140\textwidth,height=0.2\textheight,keepaspectratio]{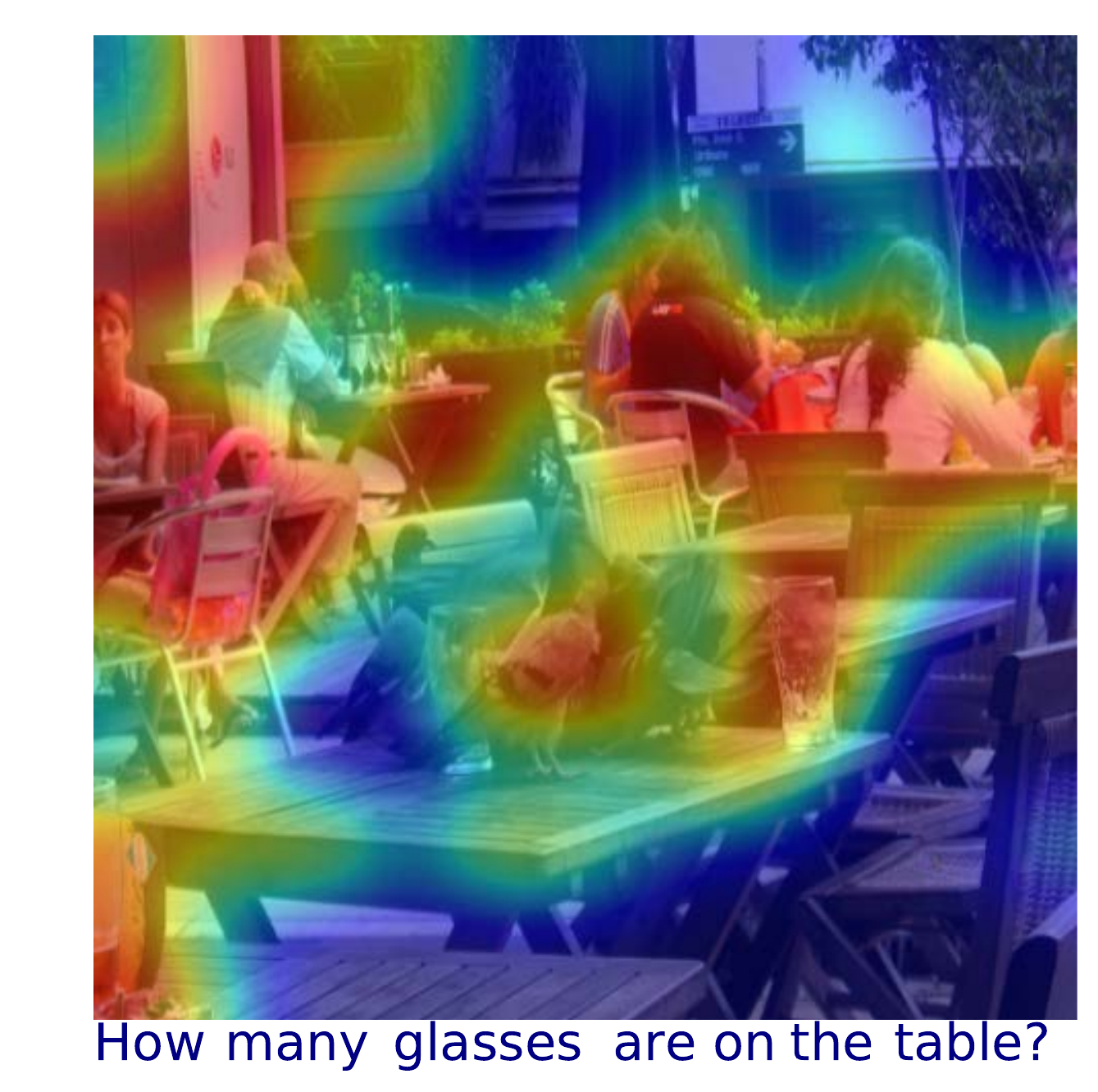} & 
		\includegraphics[clip=true,width=0.140\textwidth,height=0.2\textheight,keepaspectratio]{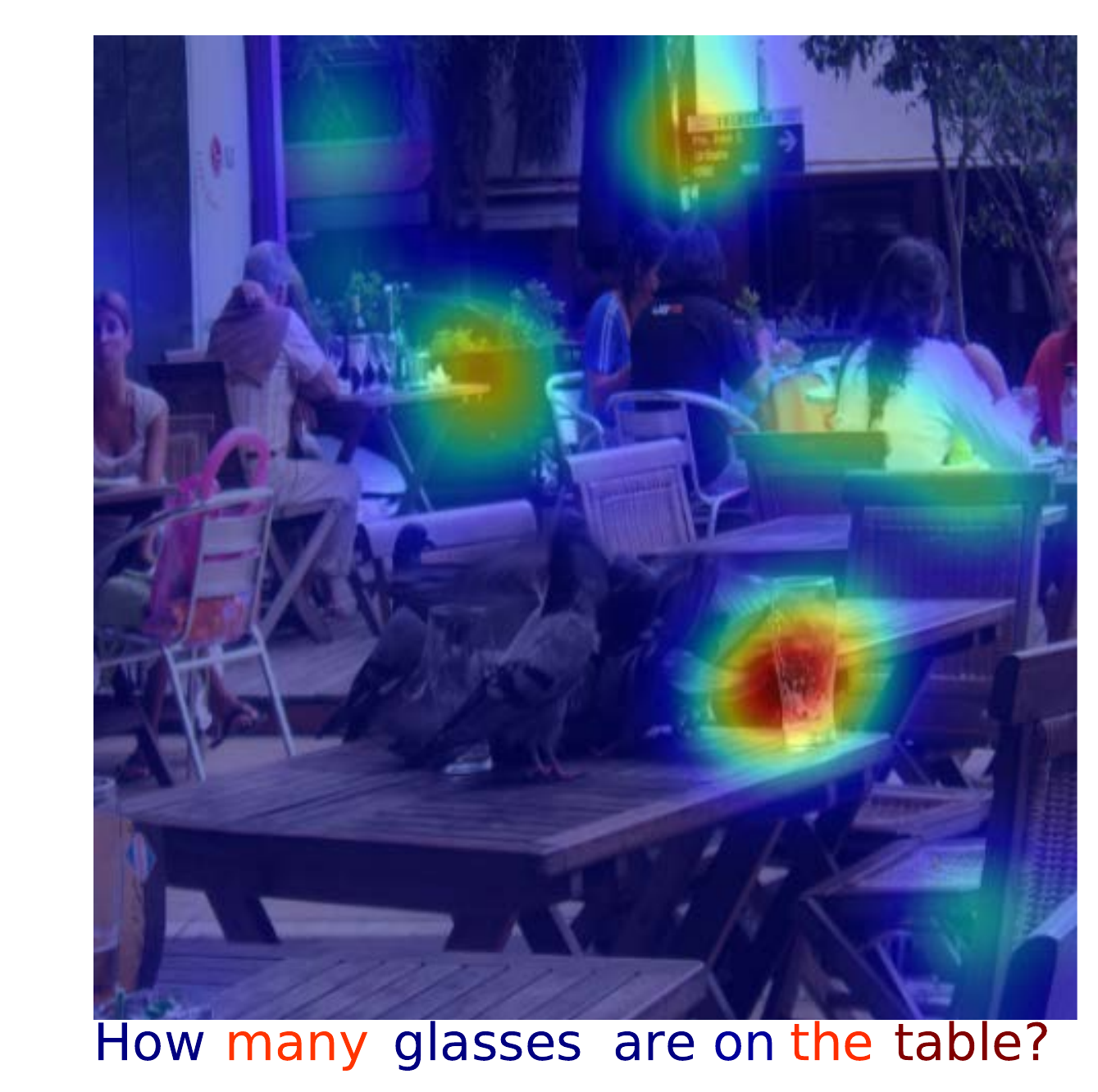} &
		\includegraphics[clip=true,width=0.140\textwidth,height=0.2\textheight,keepaspectratio]{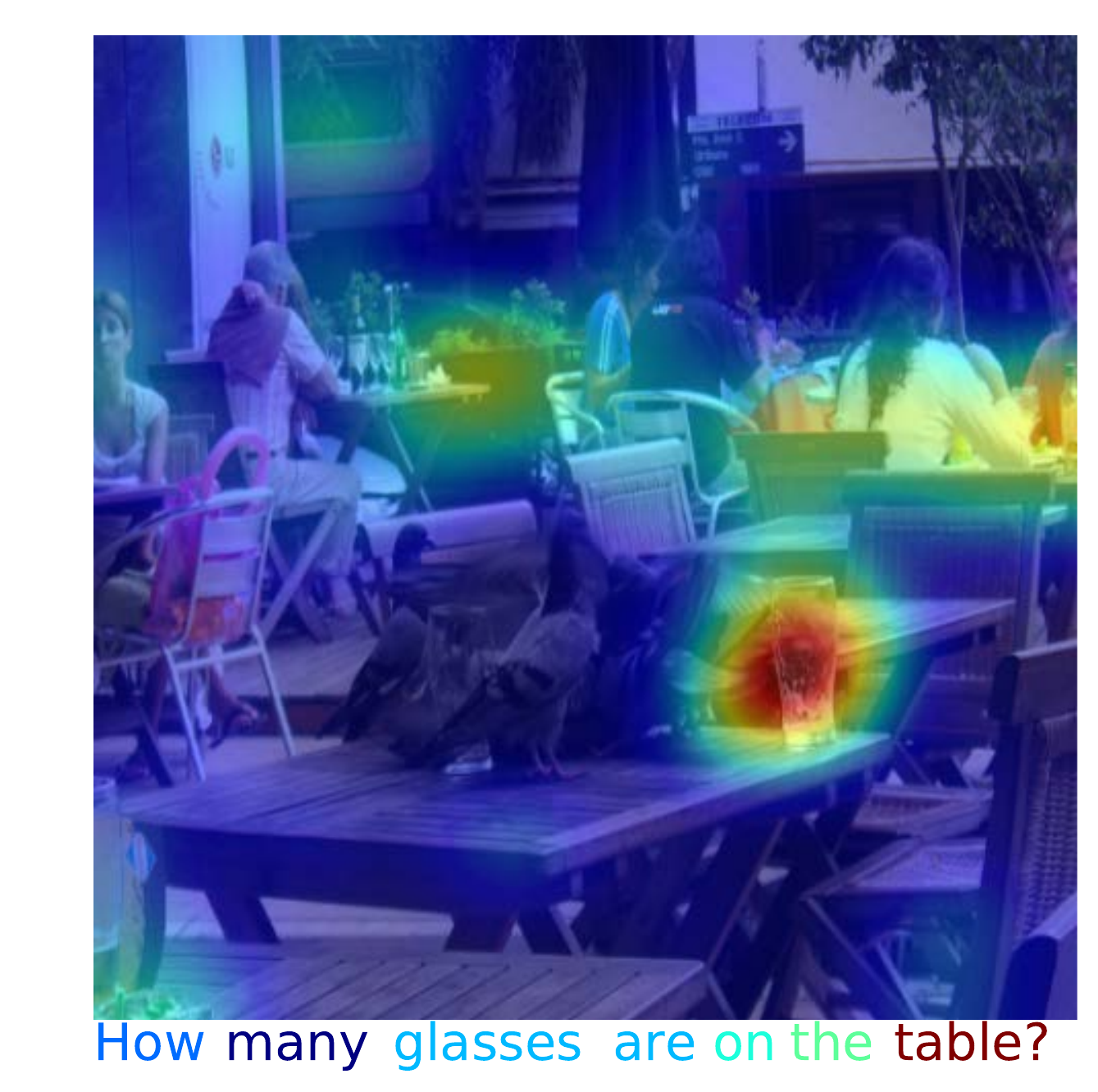} &	
		\includegraphics[clip=true,width=0.140\textwidth,height=0.2\textheight,keepaspectratio]{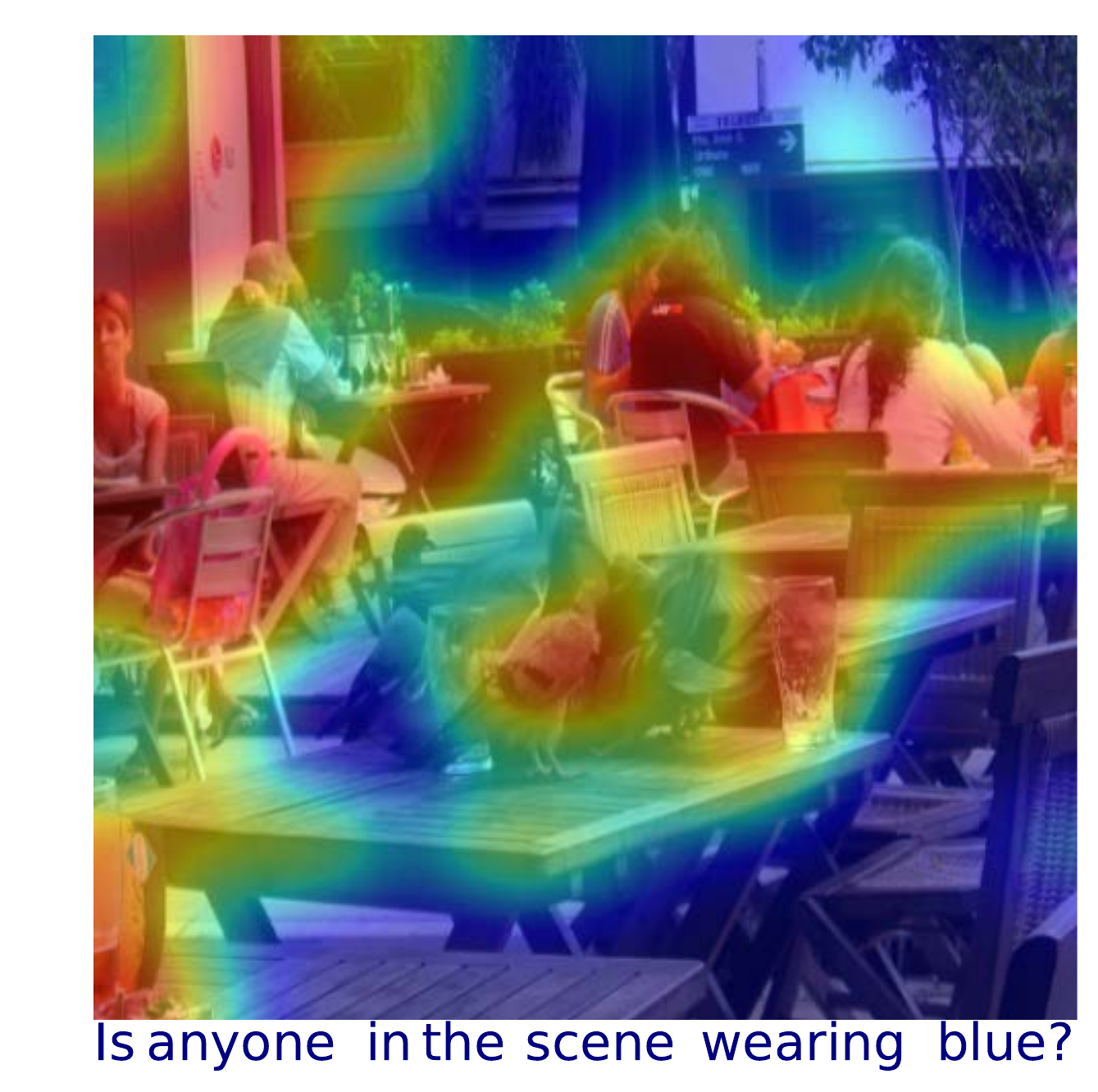} & 
		\includegraphics[clip=true,width=0.140\textwidth,height=0.2\textheight,keepaspectratio]{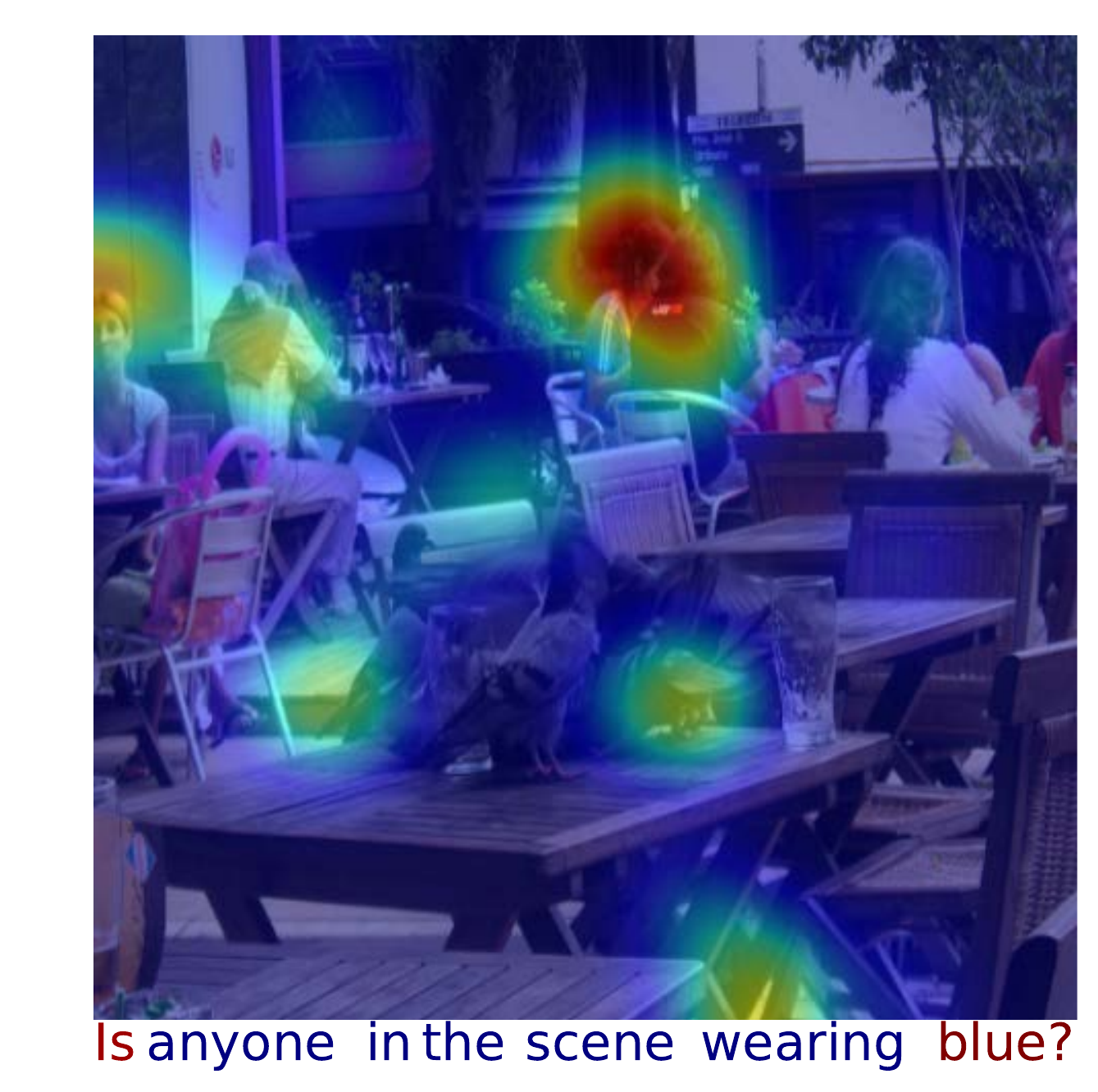} &
		\includegraphics[clip=true,width=0.140\textwidth,height=0.2\textheight,keepaspectratio]{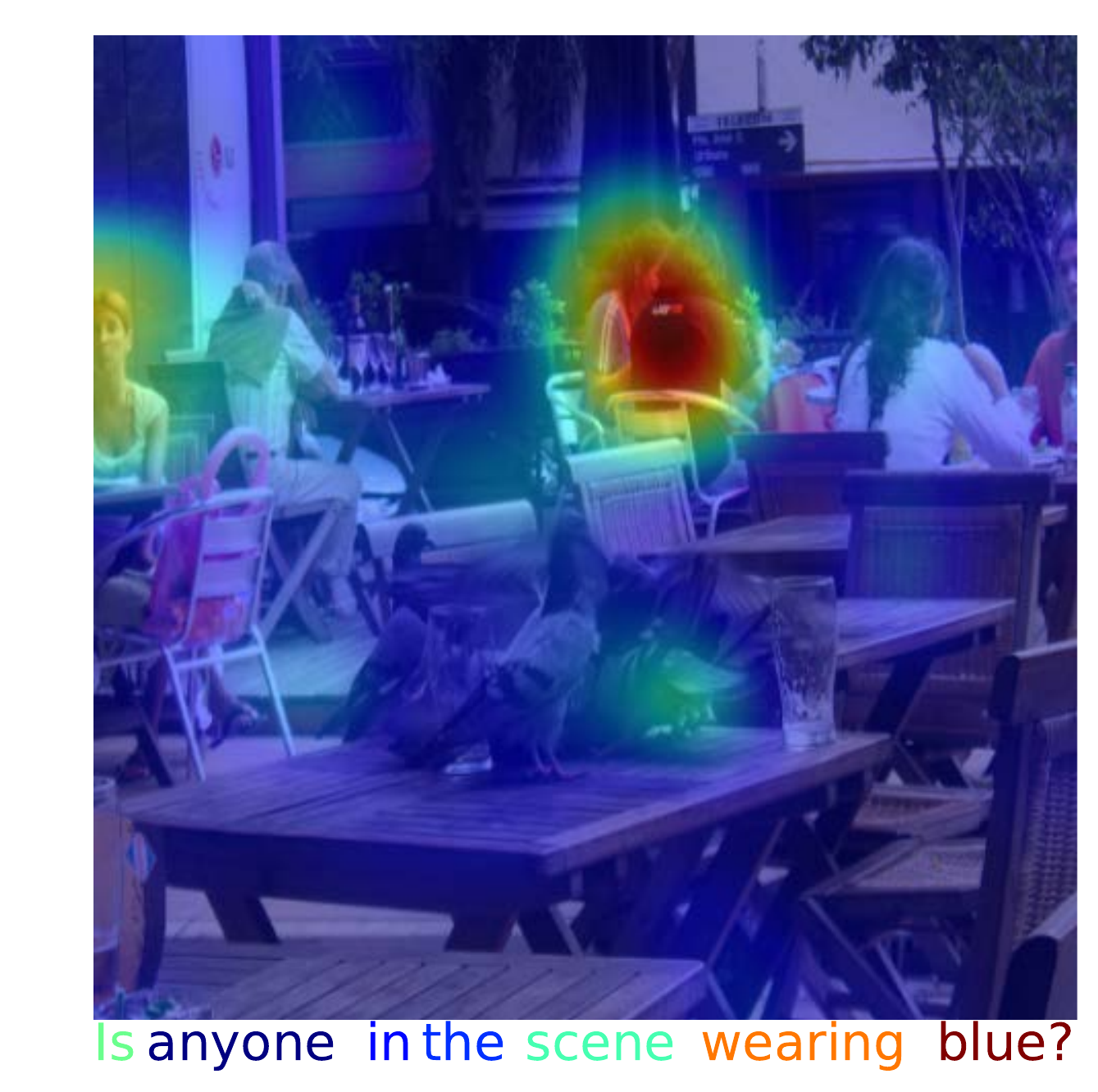} 	\\
		\raisebox{0.1cm}{\includegraphics[clip=true,width=1.8cm,height=1.8cm]{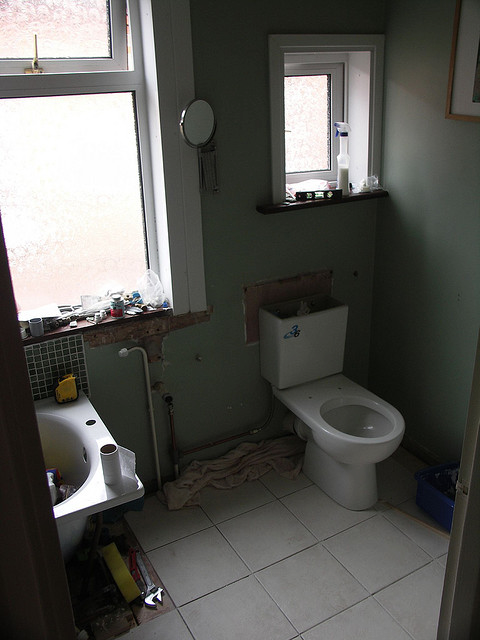}} & 
		\includegraphics[clip=true,width=0.140\textwidth,height=0.2\textheight,keepaspectratio]{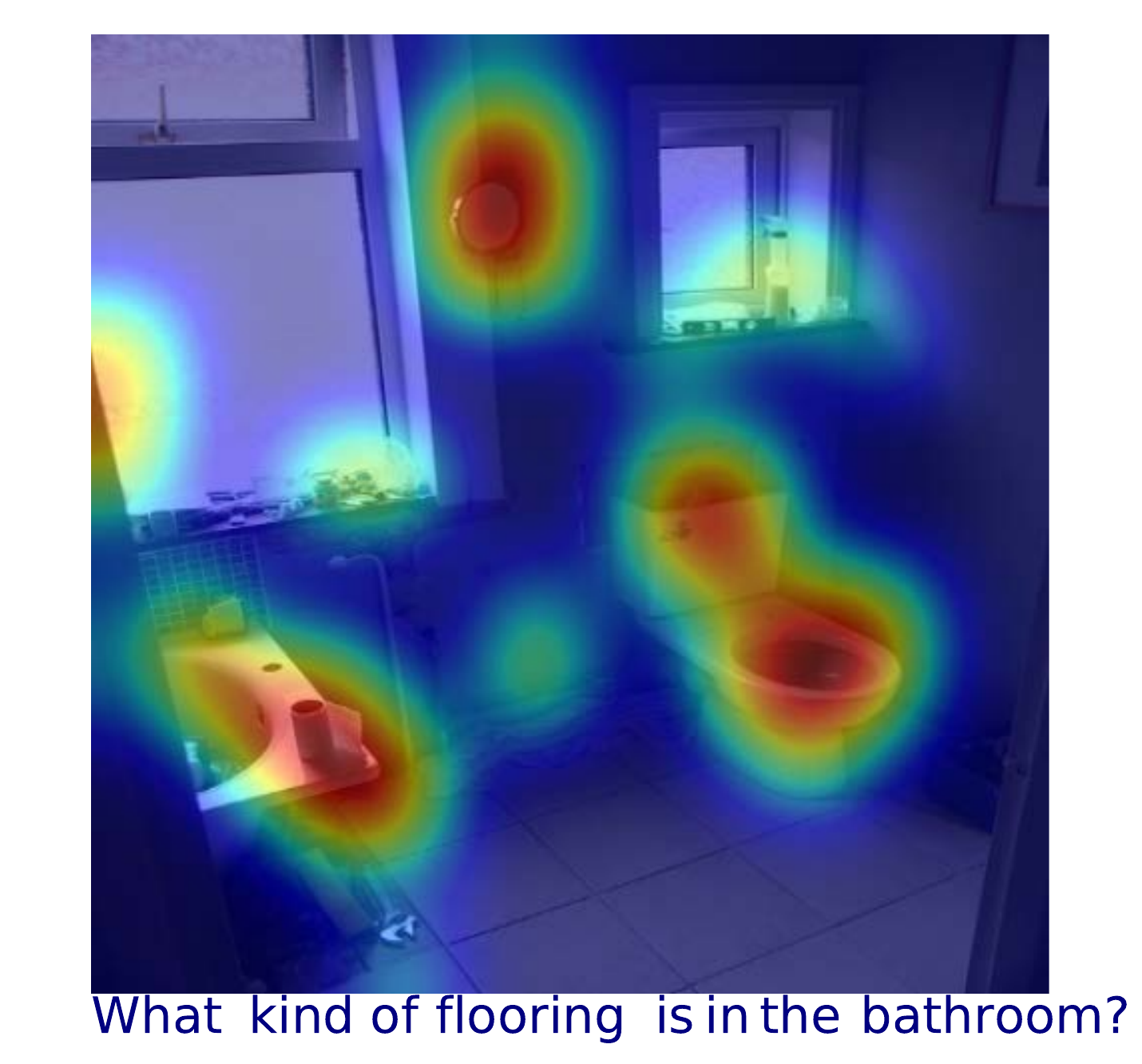} & 
		\includegraphics[clip=true,width=0.140\textwidth,height=0.2\textheight,keepaspectratio]{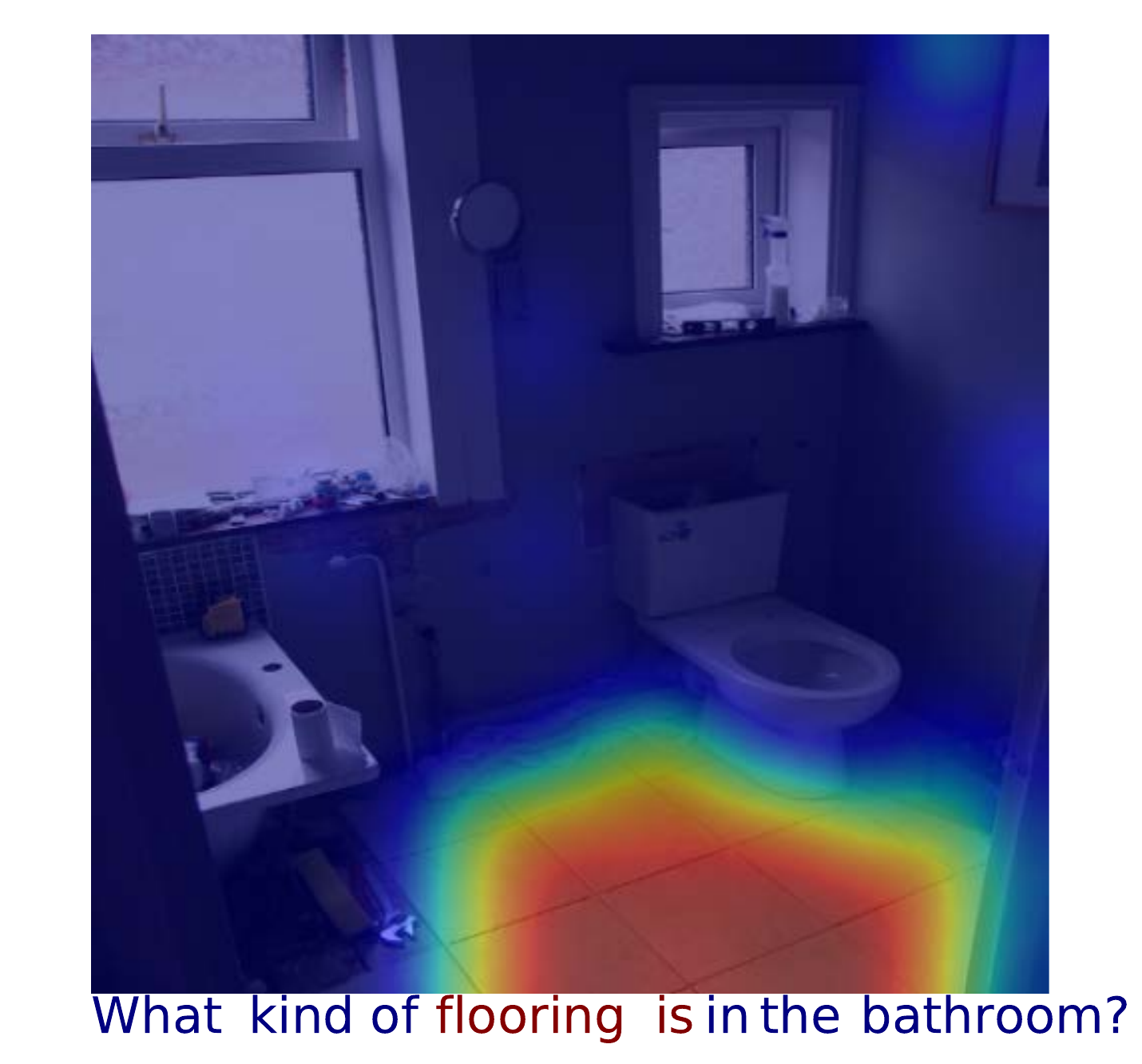} &
		\includegraphics[clip=true,width=0.140\textwidth,height=0.2\textheight,keepaspectratio]{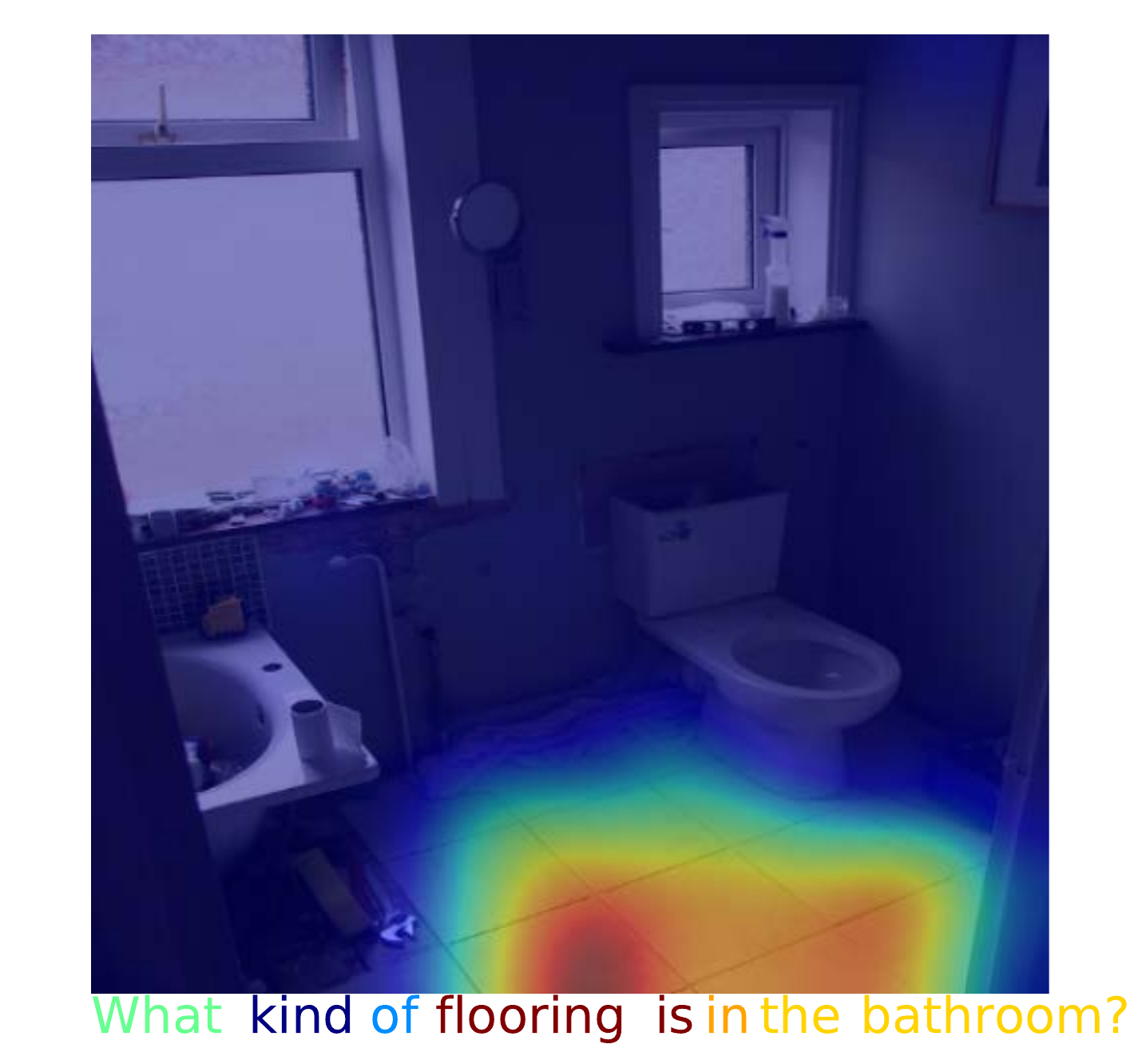} &	
		\includegraphics[clip=true,width=0.140\textwidth,height=0.2\textheight,keepaspectratio]{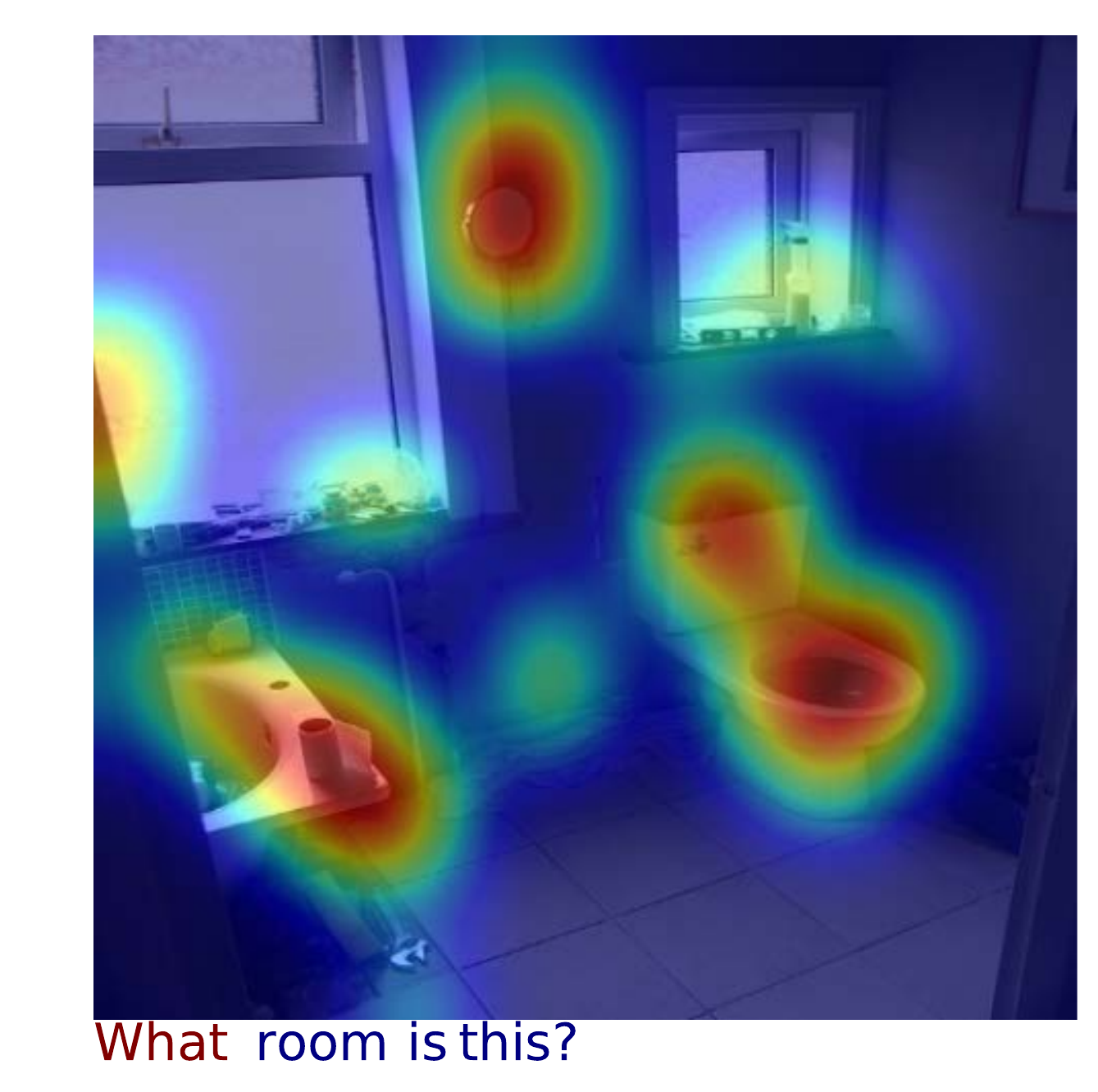} & 
		\includegraphics[clip=true,width=0.140\textwidth,height=0.2\textheight,keepaspectratio]{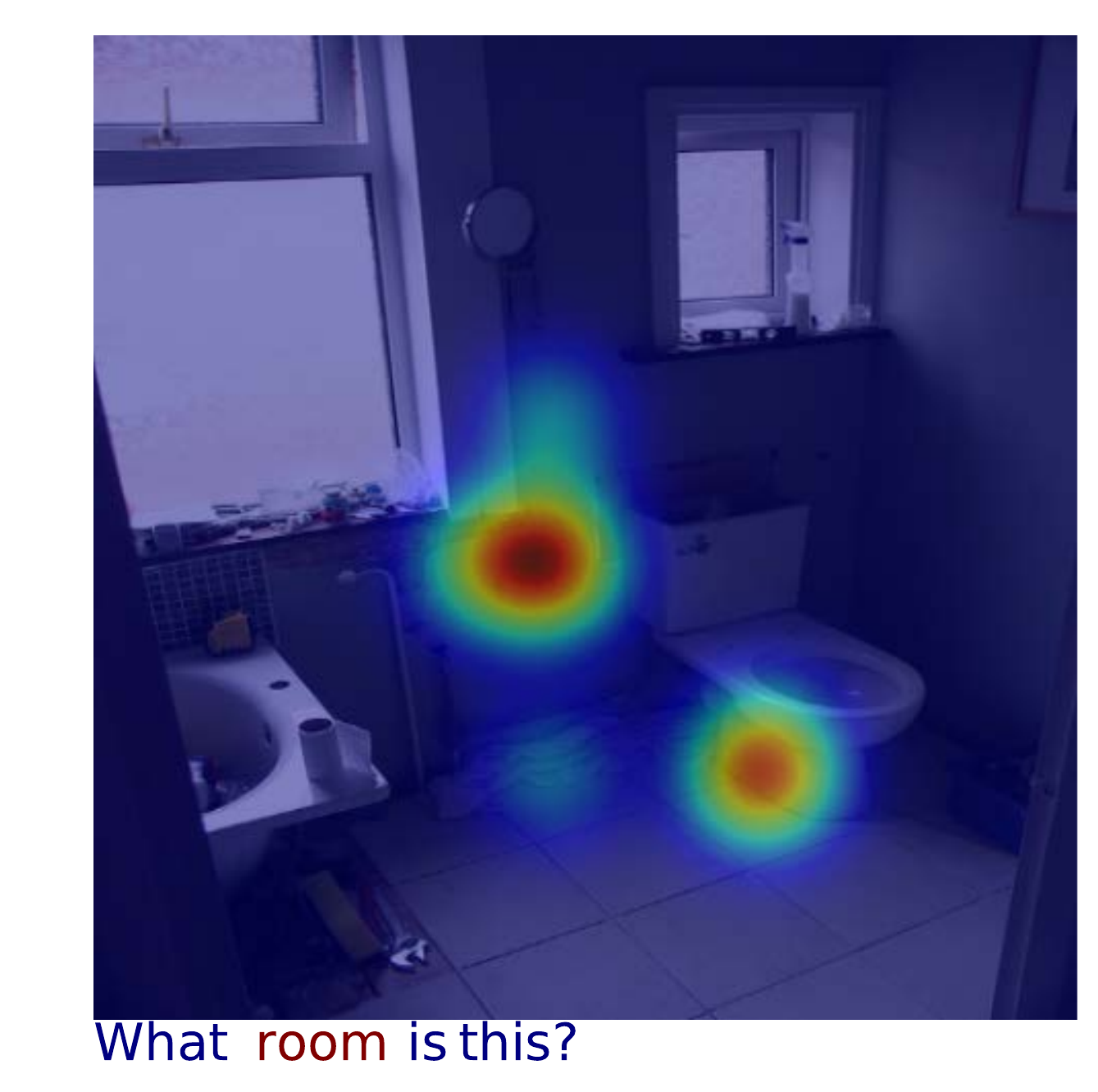} &
		\includegraphics[clip=true,width=0.140\textwidth,height=0.2\textheight,keepaspectratio]{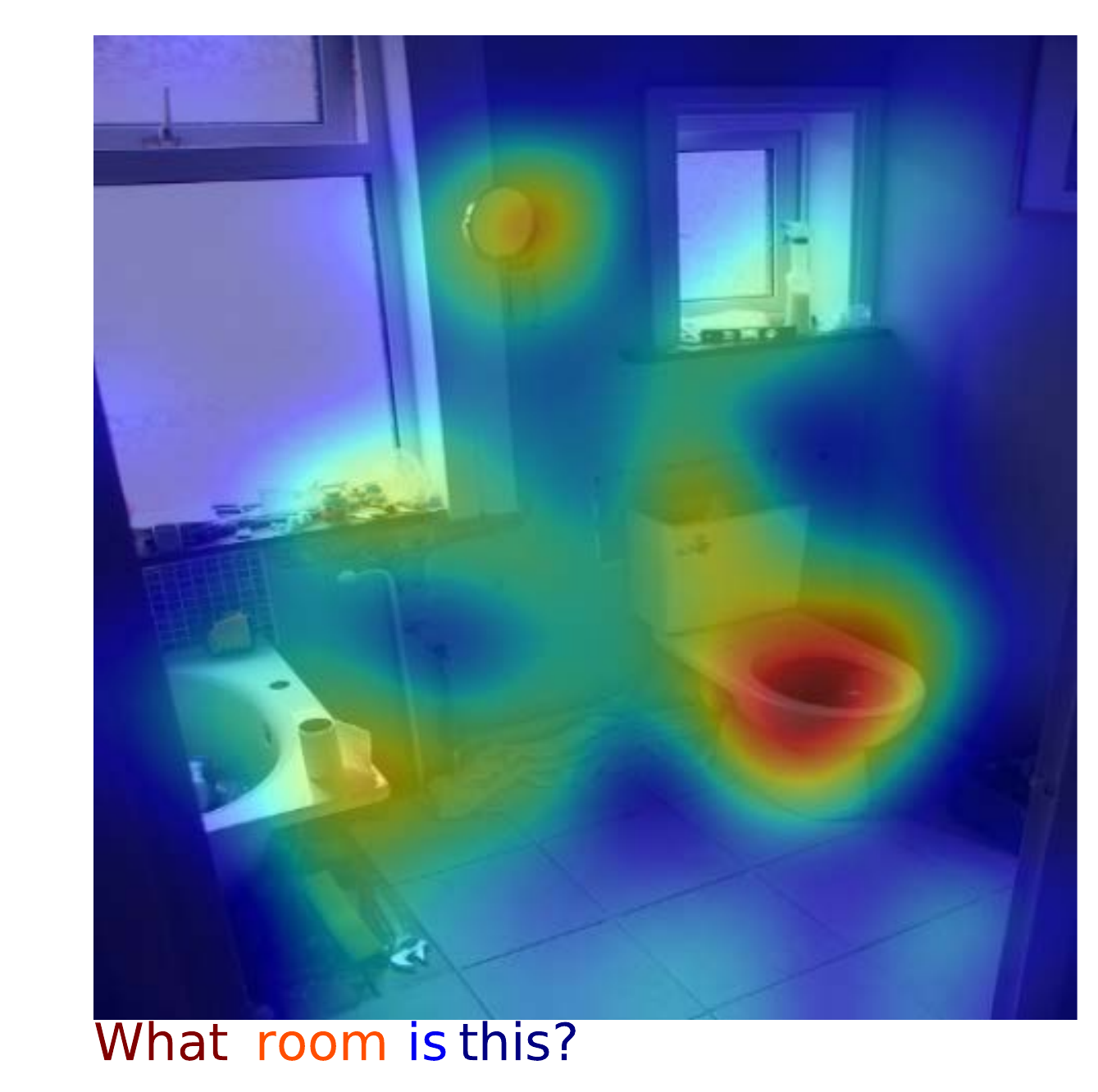} 			
	\end{tabular} 
 \vspace{-0.2cm}
	\caption{ \small
		For each image (1$^\text{st}$ column) we show the attention generated for two different questions in columns 2-4 and columns 5-7 respectively. The attentions are ordered as unary attention, pairwise attention and combined attention for both the image and the question. We observe the combined attention to significantly depend on the question.
	}
    \vspace{-0.5cm}
	\label{fig:qualitative_att}
\end{figure*}

\begin{figure*}[t]
	\centering
	\setlength\tabcolsep{0cm}
	\begin{tabular}{ccccc}
		\raisebox{0.15cm}{\includegraphics[clip=true,width=1.8cm,height=1.8cm]{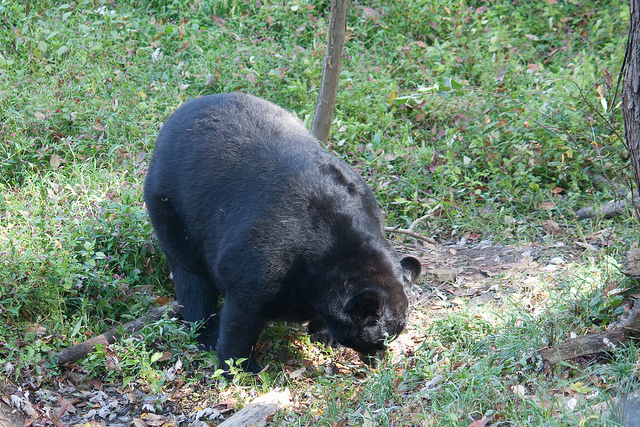}}& 
		\includegraphics[clip=true,width=0.21\textwidth,height=0.4\textheight,keepaspectratio]{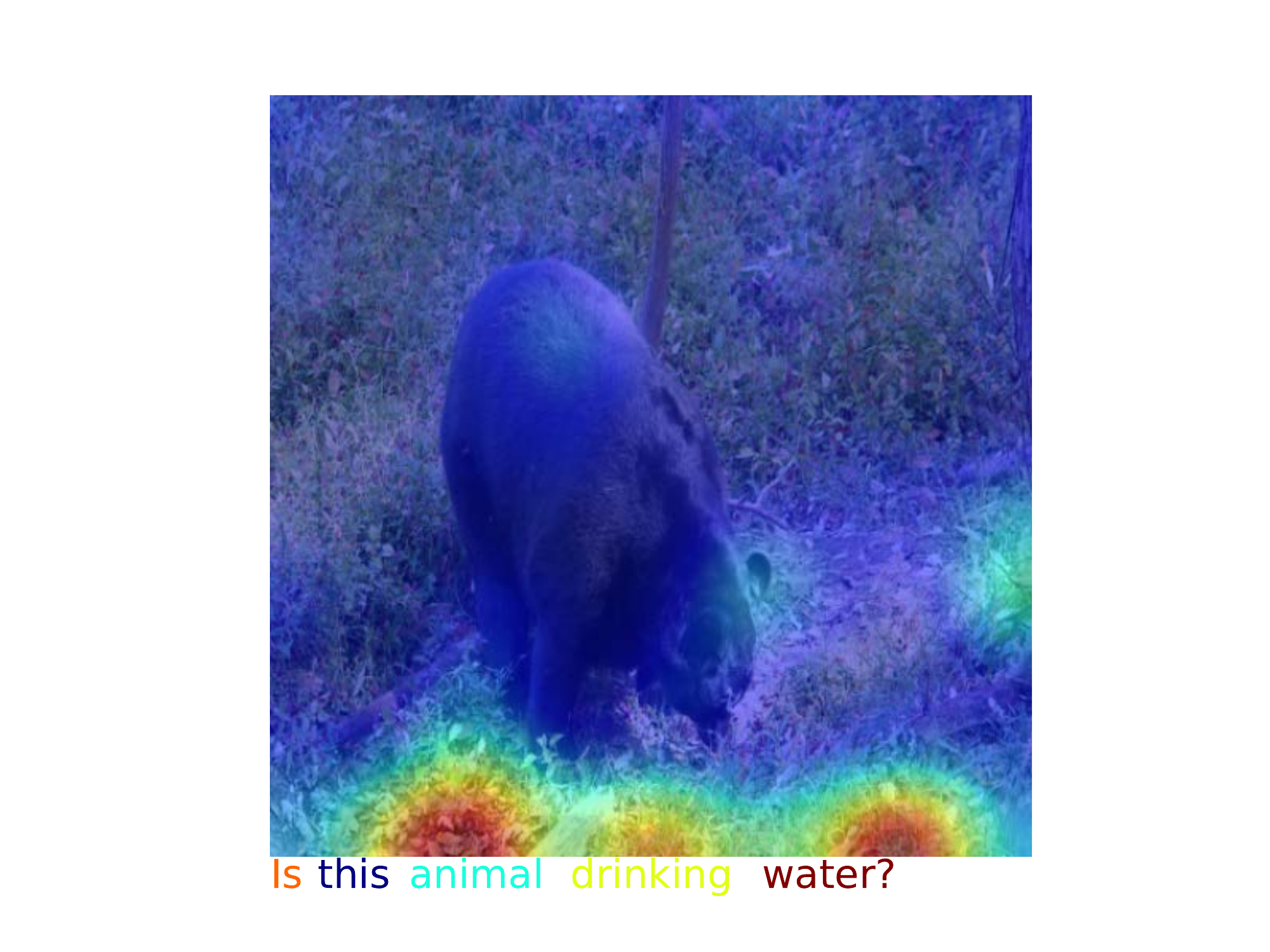} & 
		\includegraphics[clip=true,width=0.21\textwidth,height=0.4\textheight,keepaspectratio]{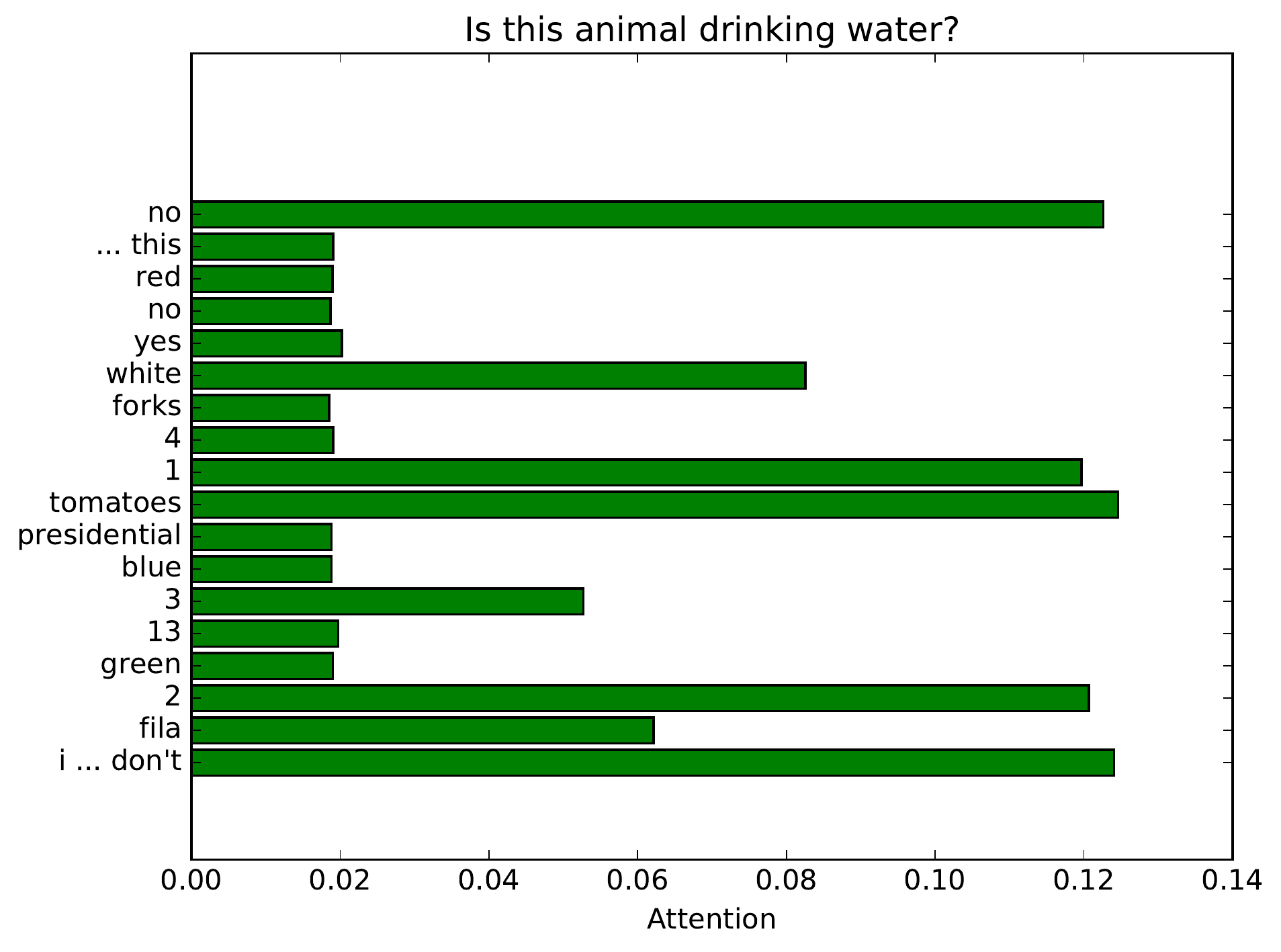} &
		\includegraphics[clip=true,width=0.21\textwidth,height=0.4\textheight,keepaspectratio]{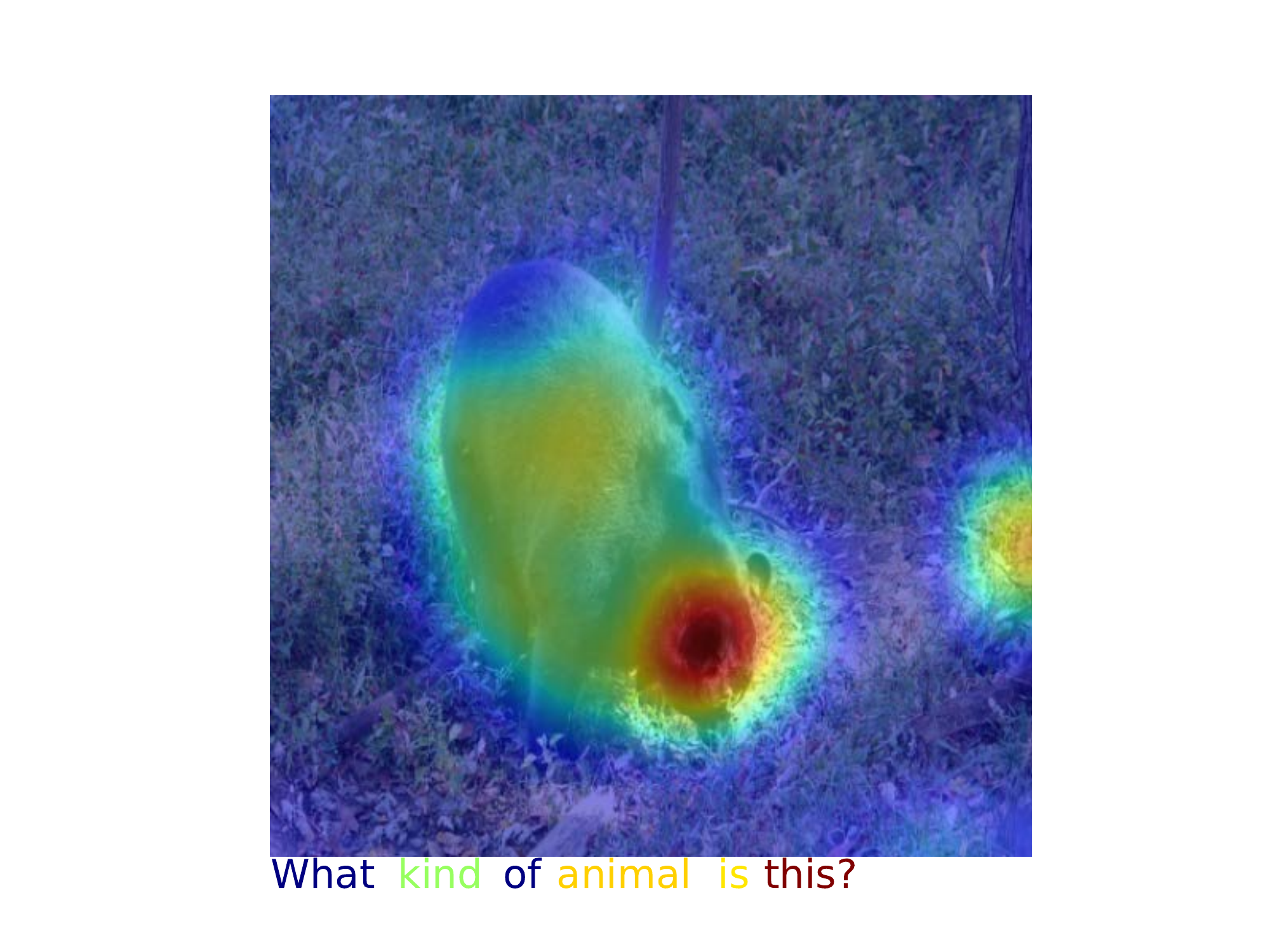} &	
		\includegraphics[clip=true,width=0.21\textwidth,height=0.4\textheight,keepaspectratio]{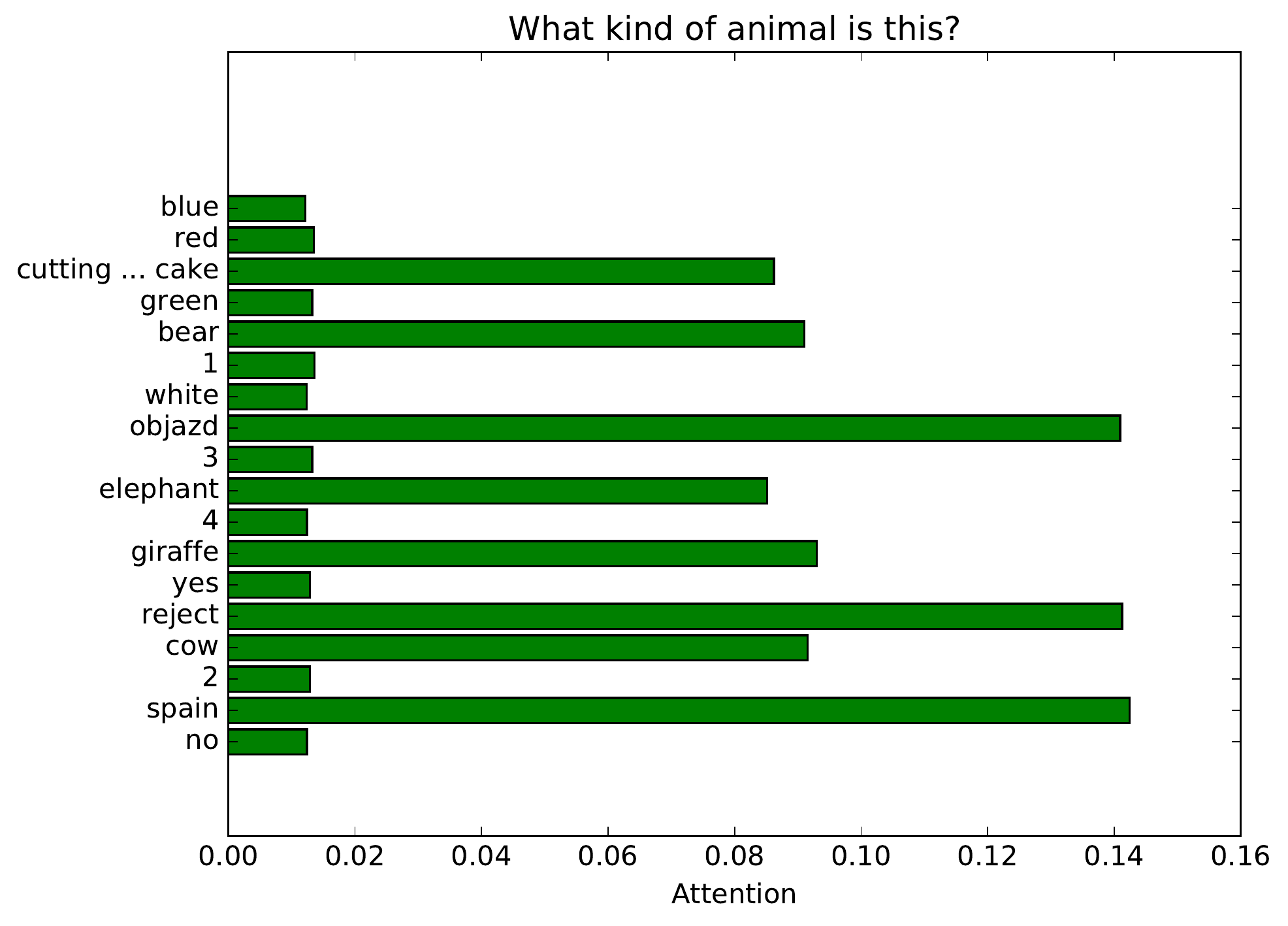} 	\\
		\raisebox{0.15cm}{\includegraphics[clip=true,width=1.8cm,height=1.8cm]{fig/3Modal/2}} & 
		\includegraphics[clip=true,width=0.21\textwidth,height=0.4\textheight,keepaspectratio]{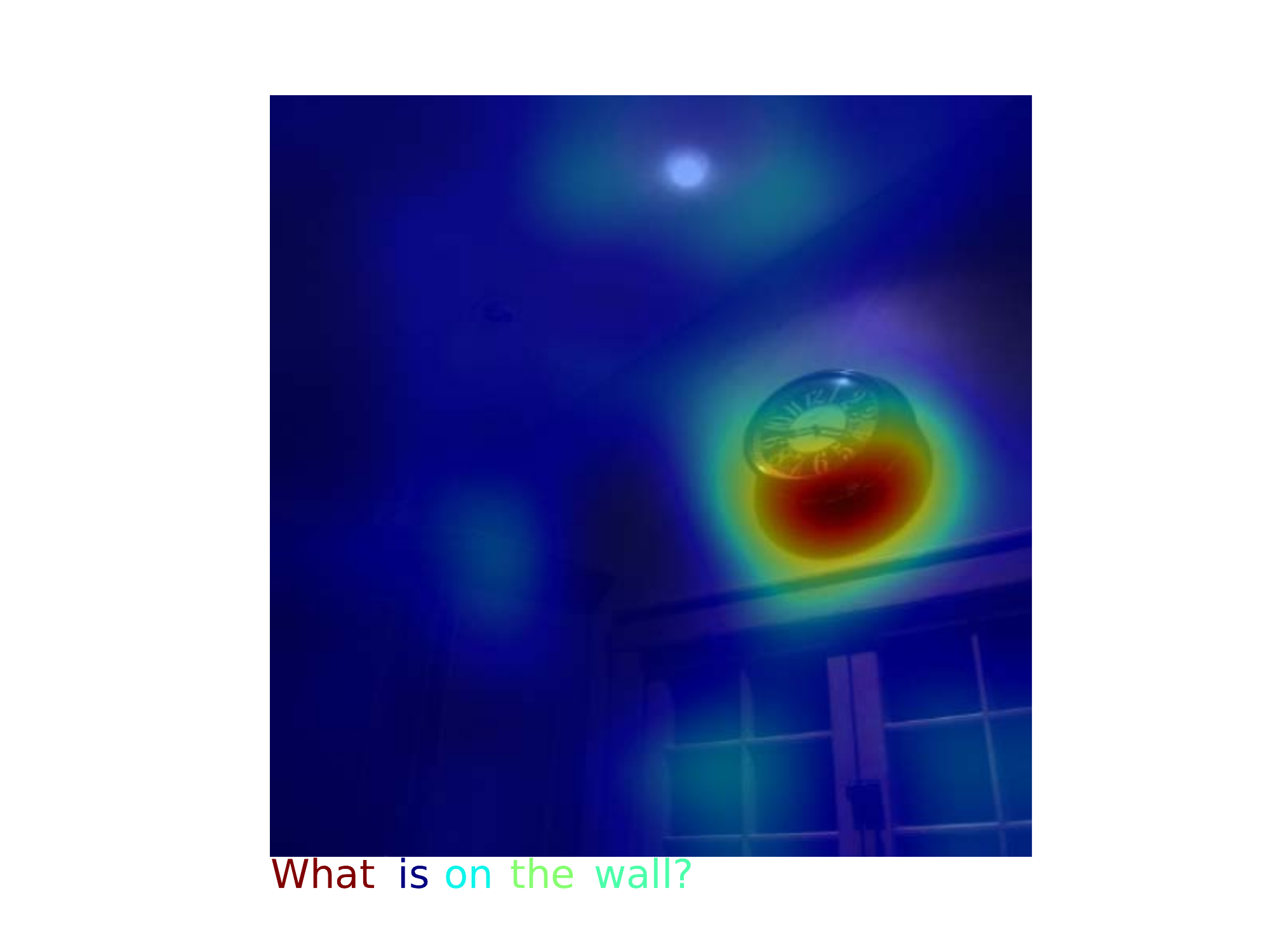} & 
		\includegraphics[clip=true,width=0.21\textwidth,height=0.2\textheight,keepaspectratio]{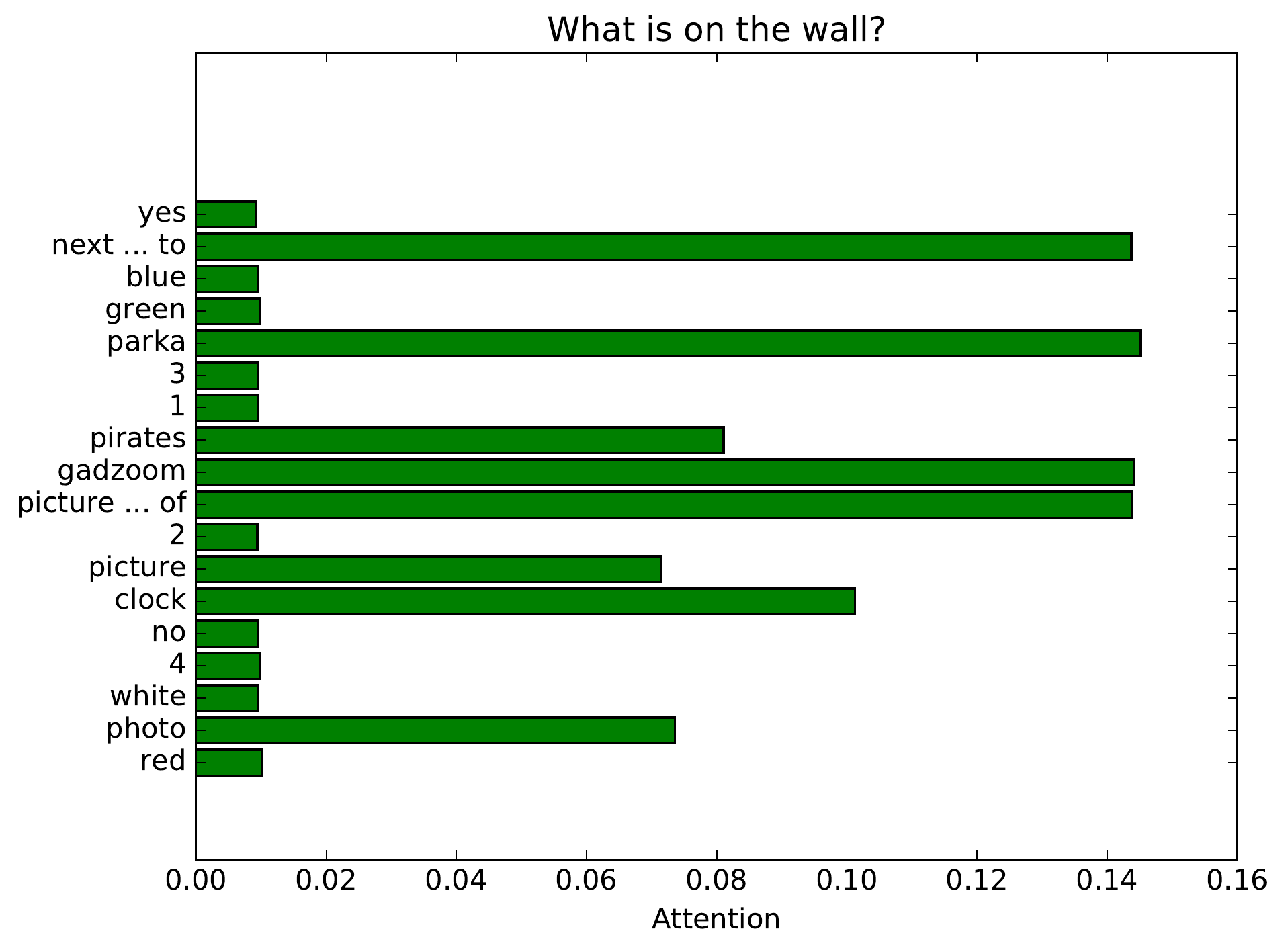} &
		\includegraphics[clip=true,width=0.21\textwidth,height=0.2\textheight,keepaspectratio]{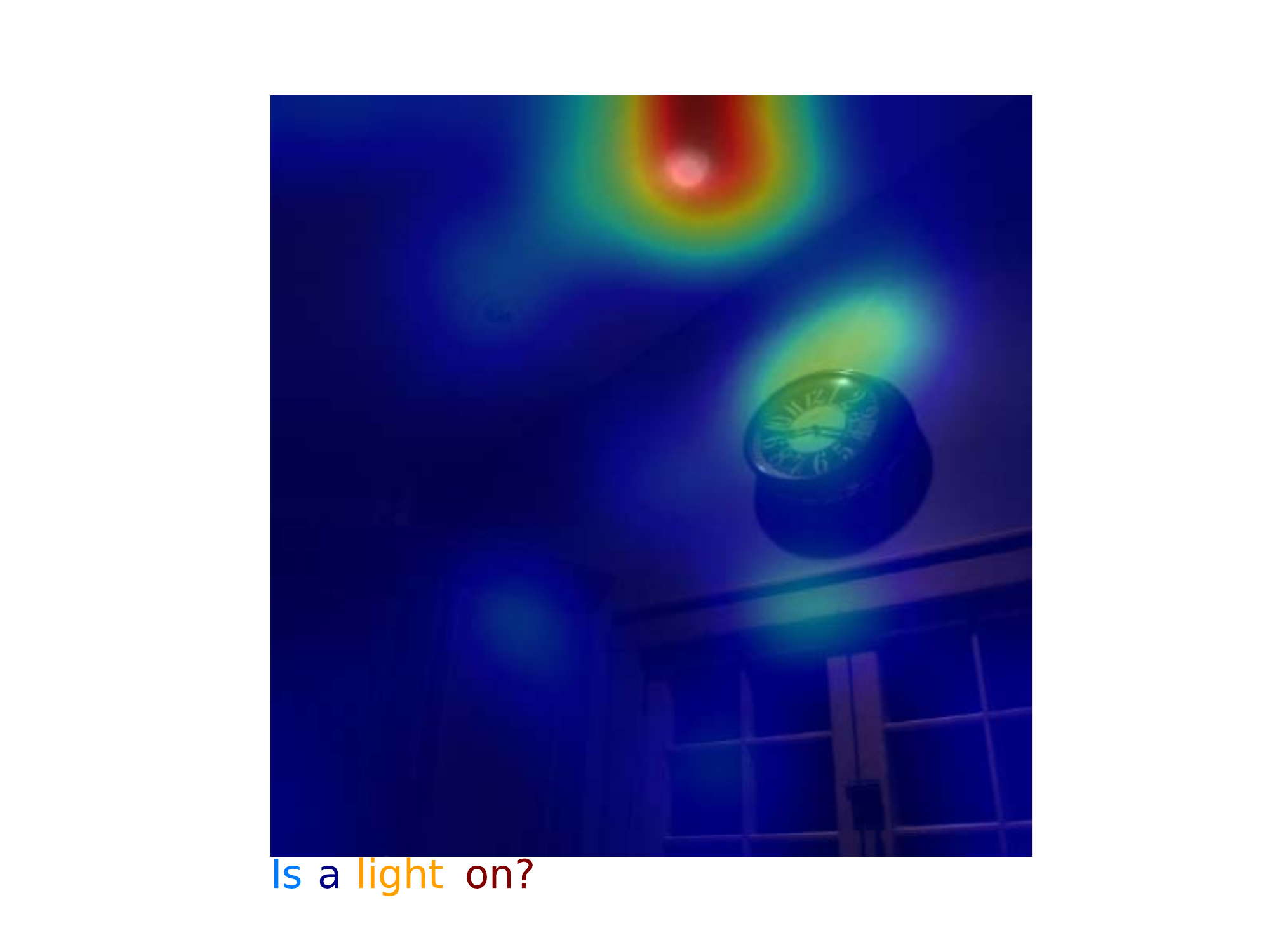} &	
		\includegraphics[clip=true,width=0.21\textwidth,height=0.2\textheight,keepaspectratio]{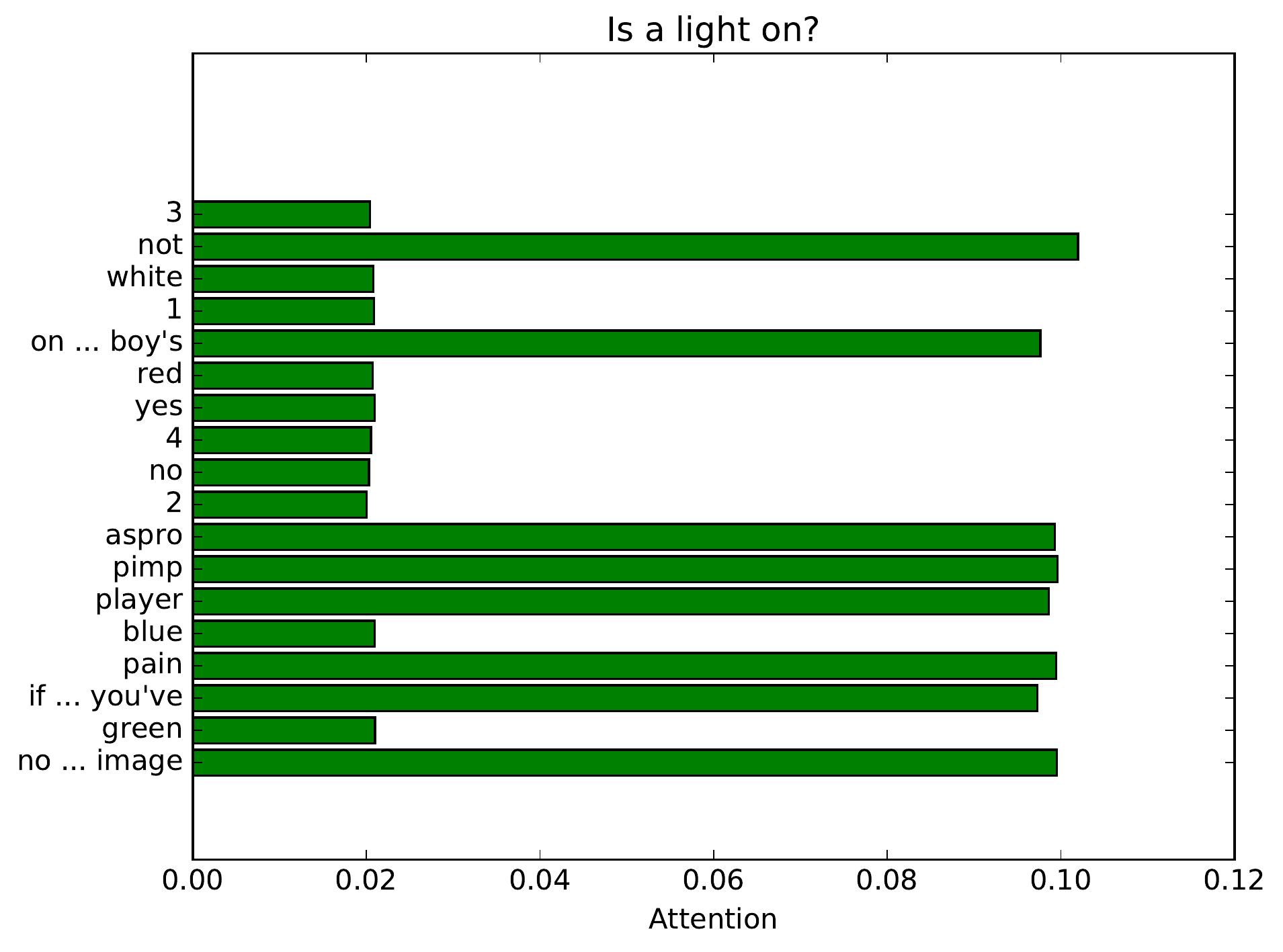}		
	\end{tabular} 
	\vspace{-0.2cm}
	\caption{ \small
		The attention generated for two different questions over three modalities.  We find the attention over multiple choice answers to emphasis the unusual answers.
	}
	\label{fig:3modal_qualtative}
\end{figure*}				
	
\begin{figure}[t]
	\centering
	\begin{tabular}{ccccc}
		\includegraphics[width=0.175\textwidth,height=65px]{fig/compare/2}& 		
		\includegraphics[width=0.175\textwidth,keepaspectratio]{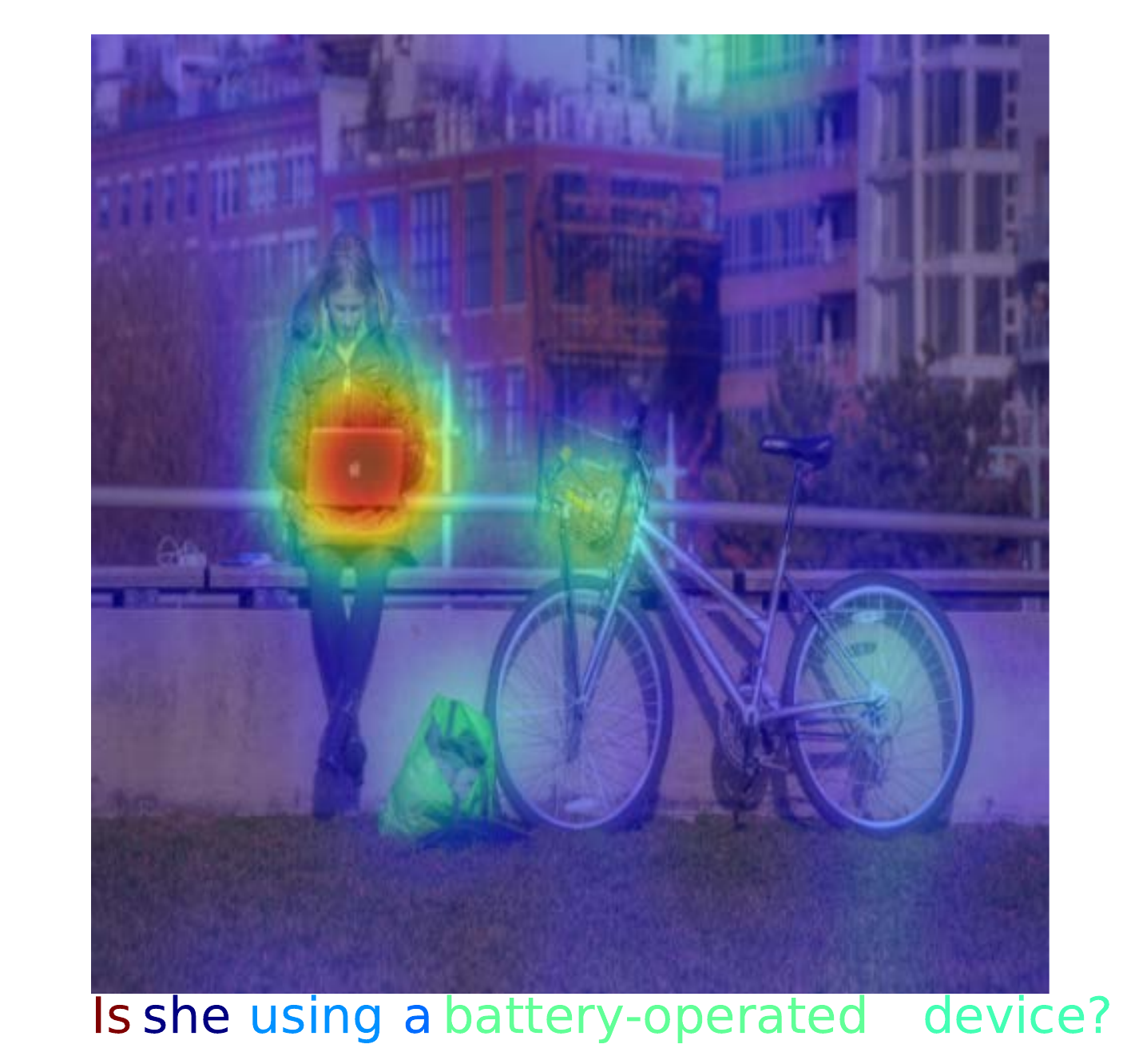}& 
		\includegraphics[width=0.175\textwidth,keepaspectratio]{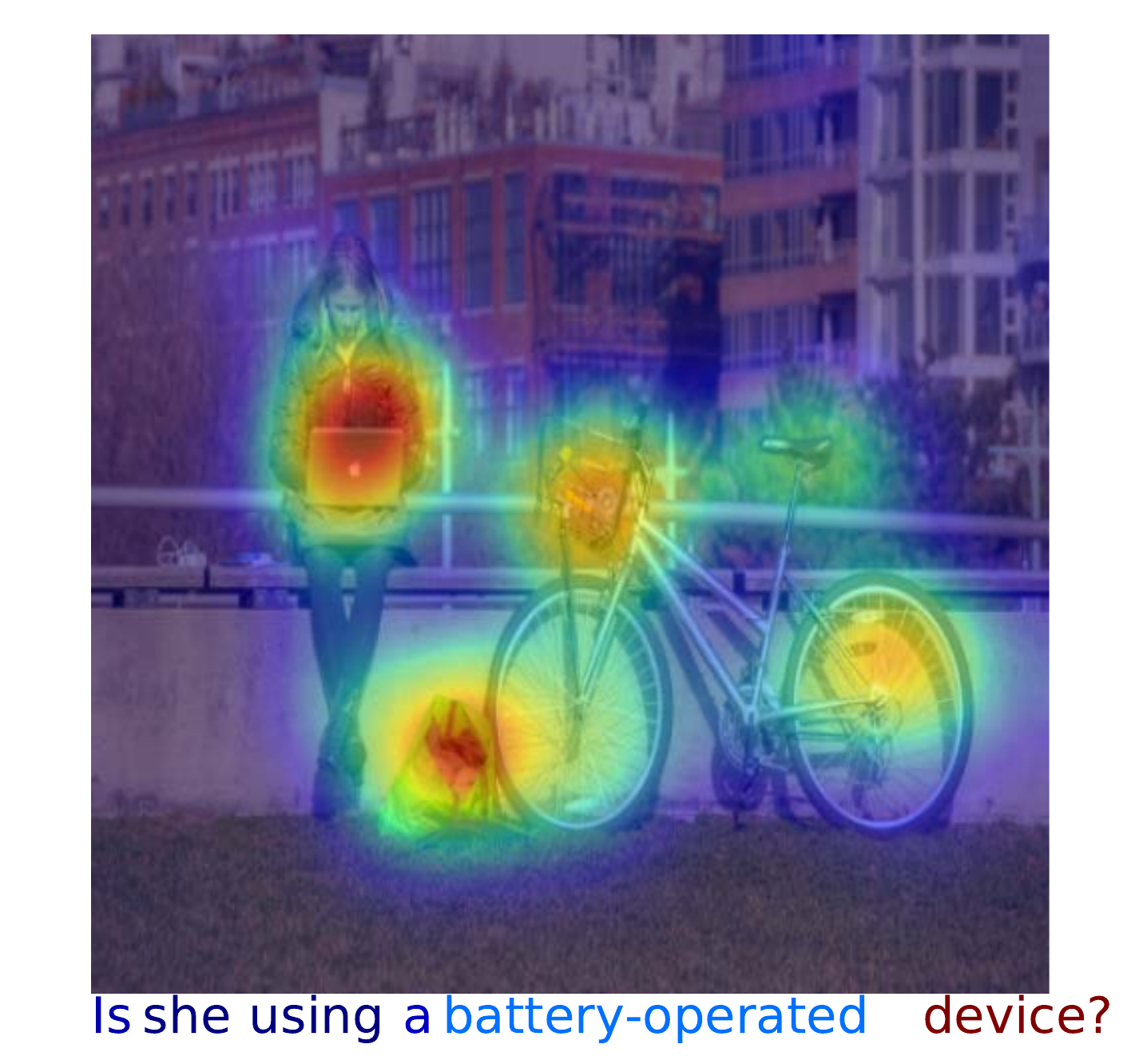}&
		\includegraphics[width=0.175\textwidth,keepaspectratio]{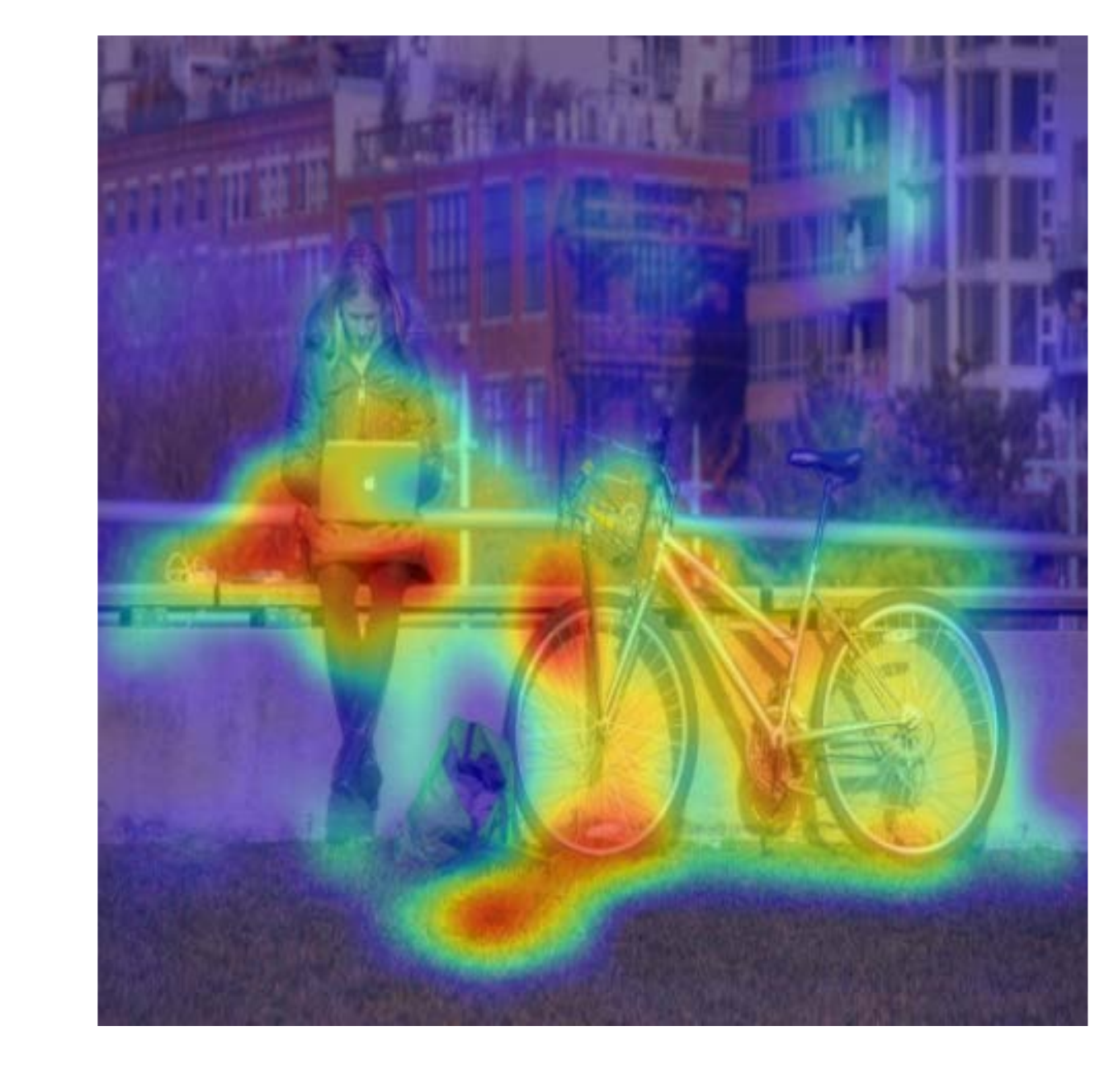}&
		\hspace{-0.4cm}\raisebox{1.08cm}{\begin{tabular}{l}
				Is she using a \\ battery device?  \\
				{\bf Ours:} yes  \\
				\cite{lu2016hierarchical}: no   \\ 
				\cite{fukui2016multimodal}: no \\
				GT: yes
				
			\end{tabular}}  \\
			{\includegraphics[width=0.175\textwidth,height=65px]{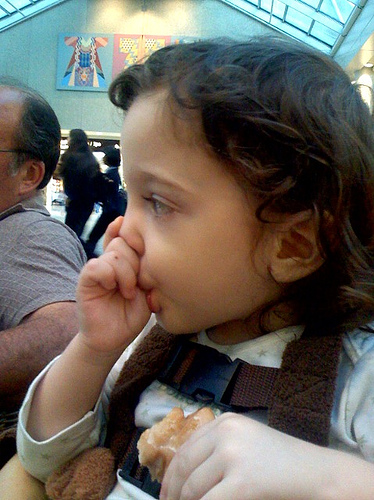} } &			
			{\includegraphics[width=0.175\textwidth,keepaspectratio]{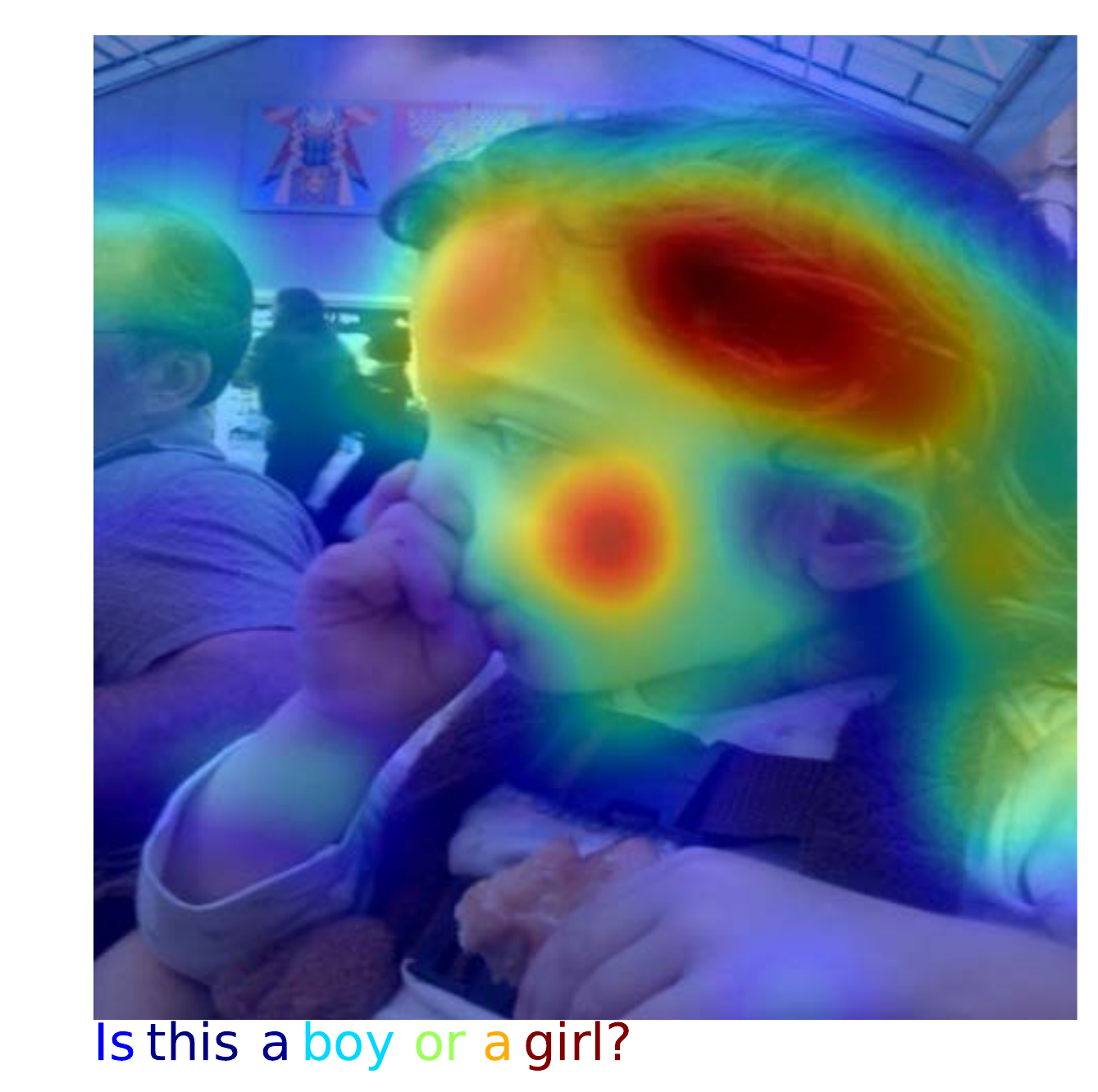} } &
			\includegraphics[width=0.175\textwidth,keepaspectratio]{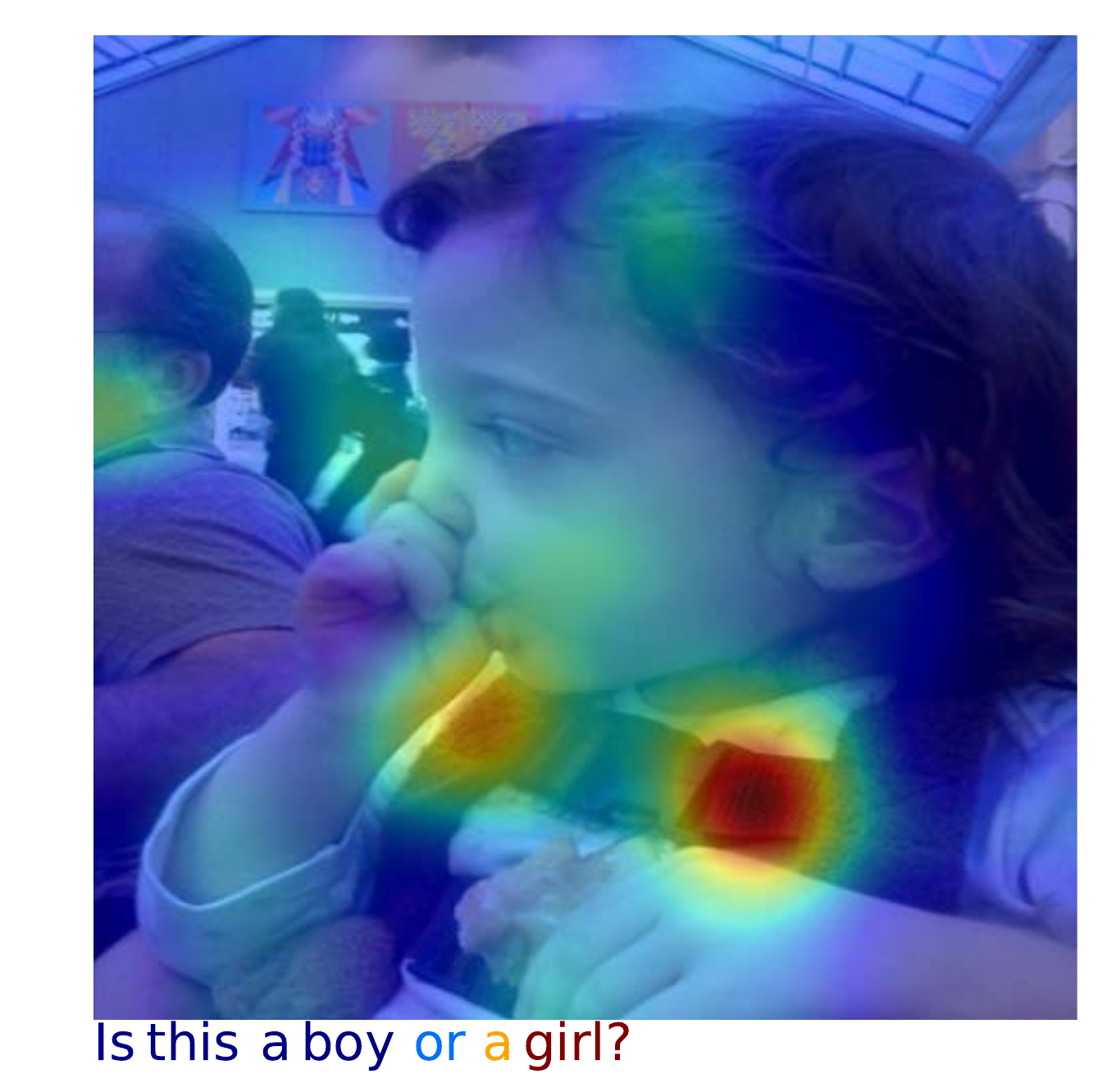}&
			\includegraphics[width=0.175\textwidth,keepaspectratio]{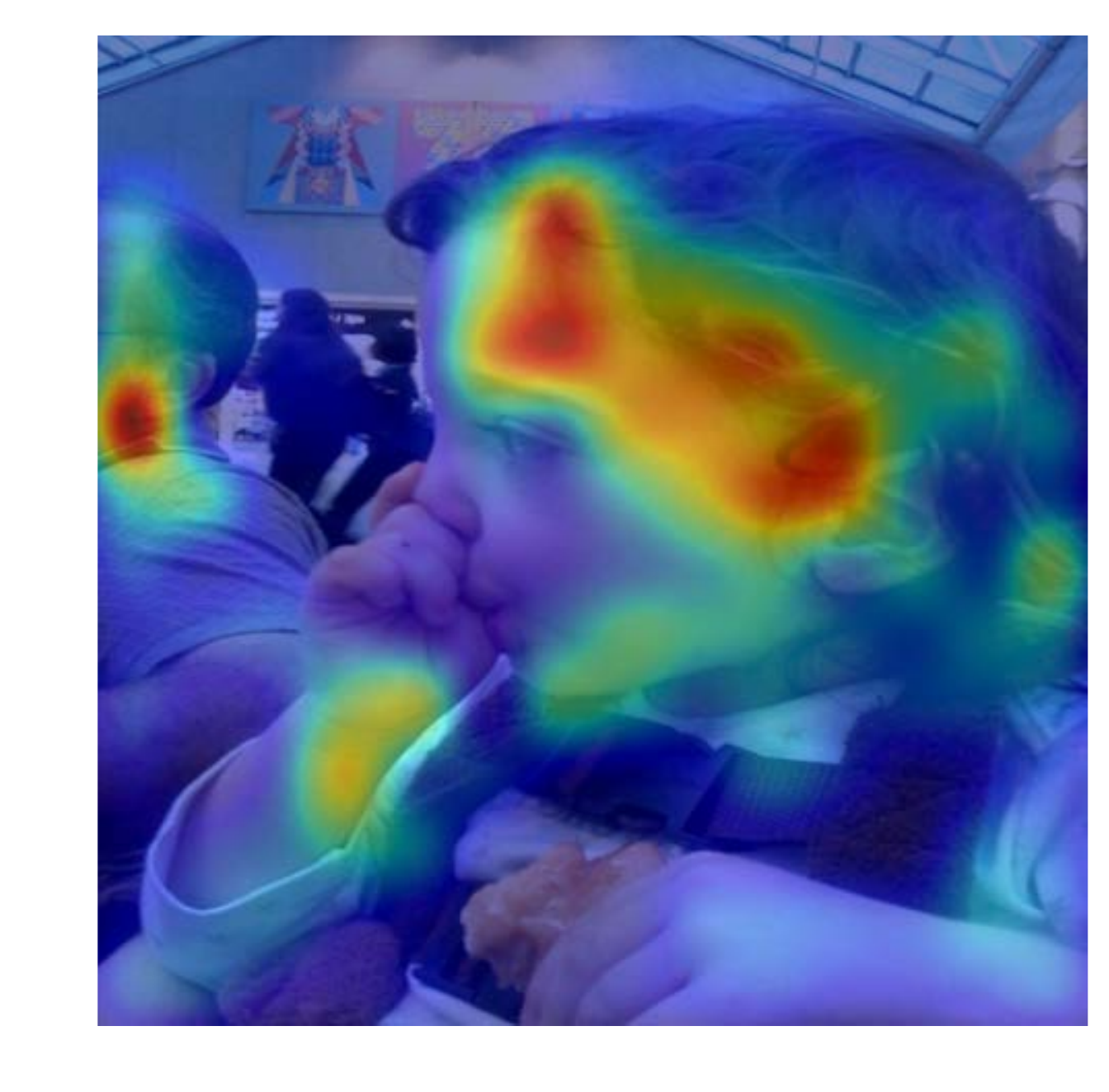}&
			\hspace{-1cm}\raisebox{1.08cm}{\begin{tabular}{l}
					Is this boy\\   or a girl? \\
					{\bf Ours:} girl  \\
					\cite{lu2016hierarchical}: boy   \\ 
					\cite{fukui2016multimodal}: girl \\
					GT: girl
					
				\end{tabular}}  	
				
			\end{tabular}
			\caption{ \small
				Comparison of our attention results (2$^{\text{nd}}$ column) with attention provided by~\cite{lu2016hierarchical} (3$^{\text{rd}}$ column) and~\cite{fukui2016multimodal} (4$^{\text{th}}$ column). The fourth column provides the question and the answer of the different techniques.
			}
			\label{fig:CompSuccess}
			\vspace{-0.3cm}
		\end{figure}

%In the following we first discuss the experimental setup before evaluating our approach qualitatively and quantitatively. We also provide an ablation analysis.

\noindent\textbf{Experimental setup:}
We use the RMSProp optimizer with a base learning rate of $4e^{-4}$ and $ \alpha = 0.99$ as well as $\epsilon = 1e^{-8}$. The batch size is set to 300. The dimension $d$ of all hidden layers is set to 512. The MCB unit feature dimension was set to $d=8192$. We apply dropout with a rate of $0.5$ after the word embeddings, the LSTM layer, and the first conv layer in the unary potential units. Additionally, for the last fully connected layer we use a dropout rate of $0.3$. We use the top 3000 most frequent answers as possible outputs, which covers 91\% of all answers in the train set. We implemented our models using the Torch framework\footnote{https://github.com/idansc/HighOrderAtten}~\cite{collobert2011torch7}. 

As a comparison for our attention mechanism we use the approach of Lu \etal~\cite{lu2016hierarchical} and the technique of Fukui \etal~\cite{fukui2016multimodal}. Their methods are based on a hierarchical attention mechanism and multi-modal compact bilinear (MCB) pooling.  In contrast to their approach we demonstrate a relatively simple technique based on a probabilistic intuition grounded on potentials. For comparative reasons only, the visualized attention  is based on two modalities: image and question. % and use the unary and pairwise potentials only.

We evaluate our attention modules on the VQA real-image  test-dev and test-std datasets~\cite{antol2015vqa}. The dataset consists of $123,287$ training images and $81,434$ test set images. Each image comes with 3 questions along with 18 multiple choice answers.  %Subsequently we first evaluate our modules quantitatively before we show qualitative results.

\noindent\textbf{Quantitative evaluation:}
We first evaluate the overall performance of  our model and compare it to a variety of baselines.  \tabref{tab:compareattention} shows the performance of our model and the baselines on the test-dev and the test-standard datasets for multiple choice (MC) questions. To obtain multiple choice results we follow common practice and use the highest scoring answer among the provided ones. Our approach (\figref{fig:VQASys}) for the multiple choice answering task achieved the reported result after 180,000 iterations, which requires about 40 hours of training on the `train+val' dataset using a TitanX GPU. Despite the fact that our model has only 40 million parameters, while techniques like~\cite{fukui2016multimodal} use over 70 million parameters, we observe state-of-the-art behavior.  Additionally, we employ a 2-modality model having a similar experimental setup. We observe a significant improvement for our 3-modality model, which shows the importance of high-order attention models. Due to the fact that we use a lower embedding dimension of 512 (similar to~\cite{lu2016hierarchical}) compared to 2048 of existing 2-modality models~\cite{kim2016hadamard,fukui2016multimodal}, the 2-modality model achieves inferior performance. We believe that higher embedding dimension and proper tuning can improve our 2-modality starting point.

Additionally, we compared our proposed decision units. MCT, which is a generic extension of MCB for 3-modalities, and 2-layers MCB which has greater expressiveness (\secref{sec:DecisionExp}). Evaluating on the 'val' dataset while training on the 'train' part using the VGG features, the MCT setup yields 63.82\% where 2-layer MCB yields 64.57\%. We also tested a different ordering of the input to the  2-modality MCB and found them to yield inferior results.

\noindent\textbf{Qualitative evaluation:} Next, we evaluate our technique qualitatively. In \figref{fig:qualitative_att} we 
illustrate the unary, pairwise and combined attention of our approach based on the two modality architecture, without the multiple choice as input. For each image we show multiple questions.  
We observe the unary attention usually attends to strong features of the image, while pairwise potentials emphasize areas that correlate with question words. Importantly, the combined result is dependent on the provided question. 
For instance, in the first row we observe for the question ``How many glasses are on the table?,'' that the pairwise potential reacts to the image area depicting the glass. In contrast, for the question ``Is anyone in the scene wearing blue?'' the pairwise potentials reacts to the guy with the blue shirt. In \figref{fig:3modal_qualtative}, we illustrate the attention for our 3-modality model. We find the attention over multiple choice answers to favor the more unusual results.

In \figref{fig:CompSuccess}, we compare the final attention obtained from our approach to the results obtained with techniques discussed in~\cite{lu2016hierarchical} and~\cite{fukui2016multimodal}. We observe that our approach attends to reasonable pixel and question locations. For example, considering the first row in \figref{fig:CompSuccess}, the question refers to the battery operated device. Compared to existing approaches, our technique attends to the laptop, which seems to help in choosing the correct answer. 
In the second row, the question wonders ``Is this a boy or a girl?''. Both of the correct answers were produced when the attention focuses on the hair.  %, 

\begin{figure}[t]

\centering
		\begin{tabular}{ccccc}
			\includegraphics[width=0.175\textwidth,height=65px]{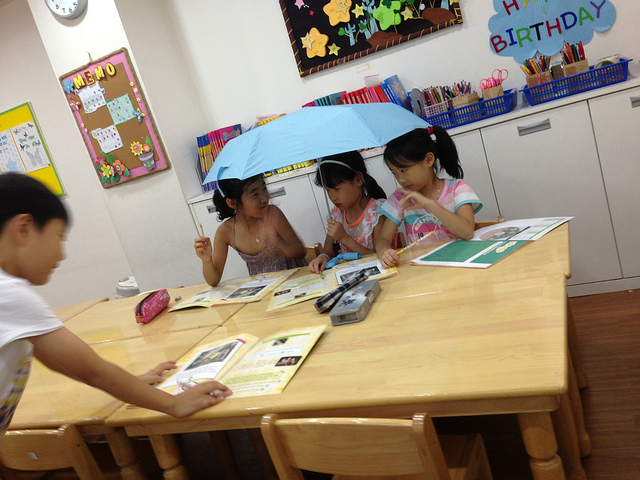} &
			\includegraphics[width=0.175\textwidth,height=65px]{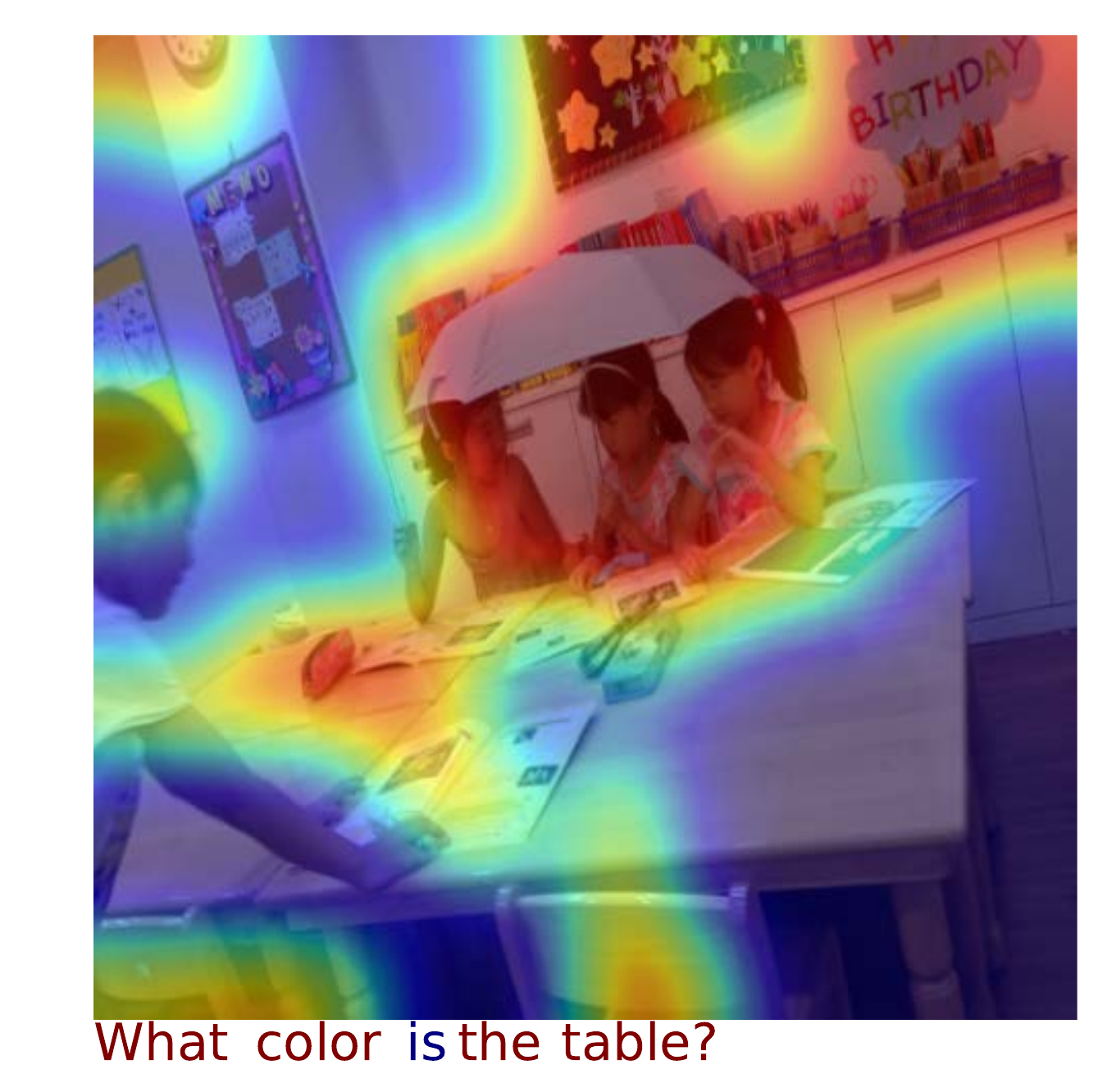} &		\includegraphics[width=0.175\textwidth,height=65px]{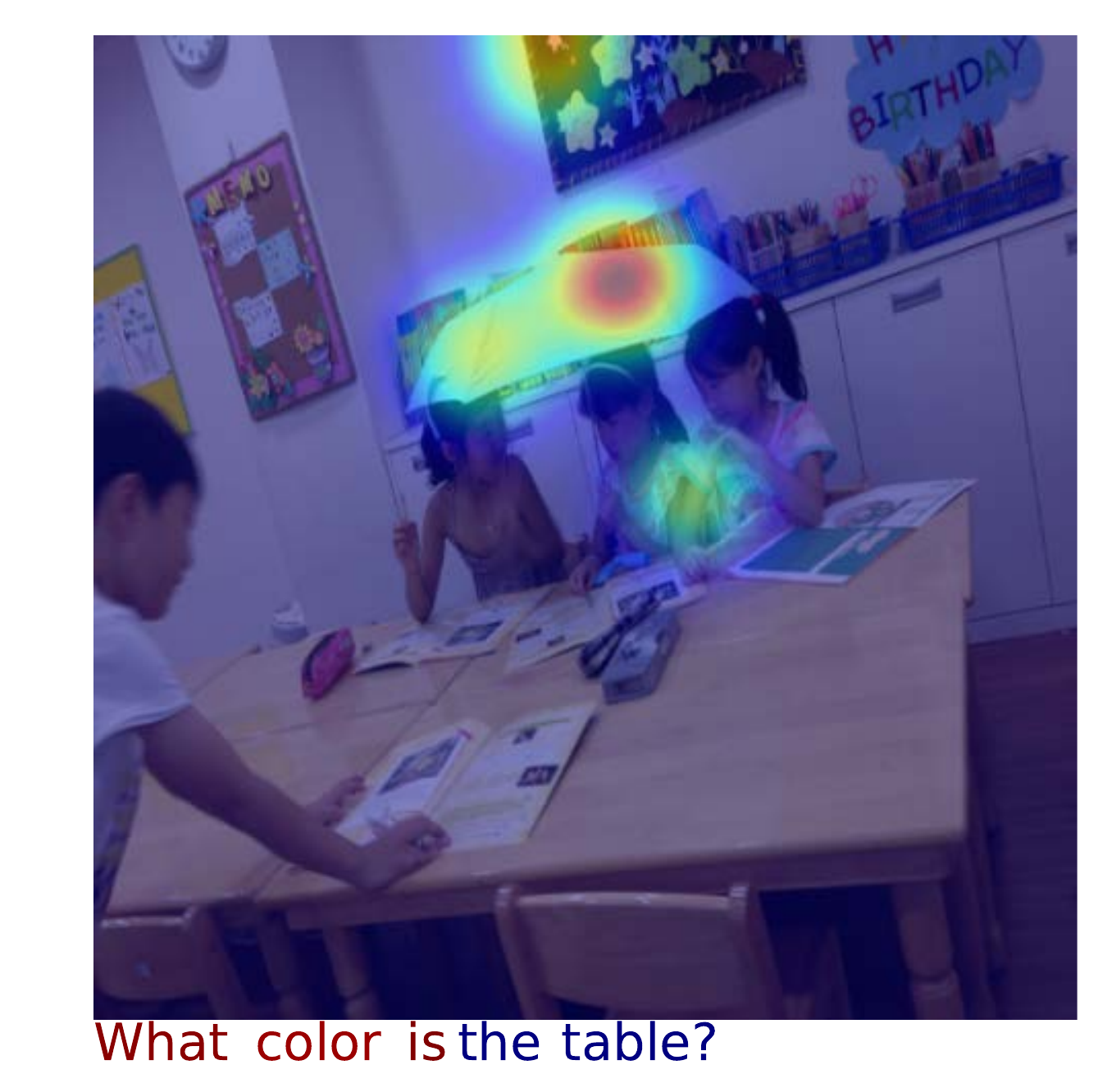} & 		\includegraphics[width=0.175\textwidth,height=65px]{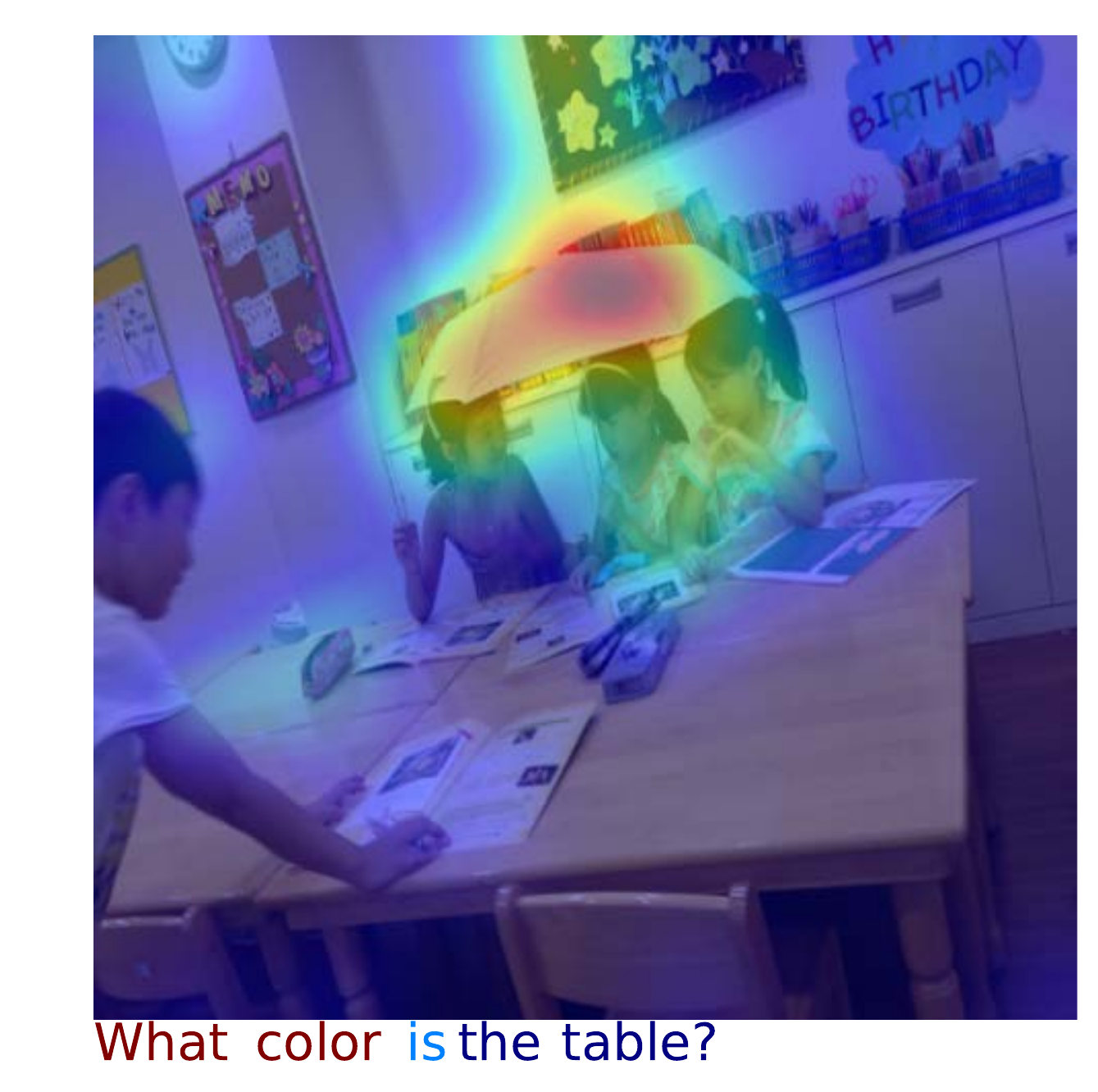} & 
			\hspace{-0.5cm}\raisebox{1.3cm}{
				{\small{
						\begin{tabular}{l}
							What color is\\ the table? \\
							GT: brown\\
							{\bf Ours:} blue
						\end{tabular}}}}
						\\ 	
			\includegraphics[width=0.175\textwidth,height=65px]{fig/atten_fail/1} &							
			\includegraphics[width=0.175\textwidth,height=65px]{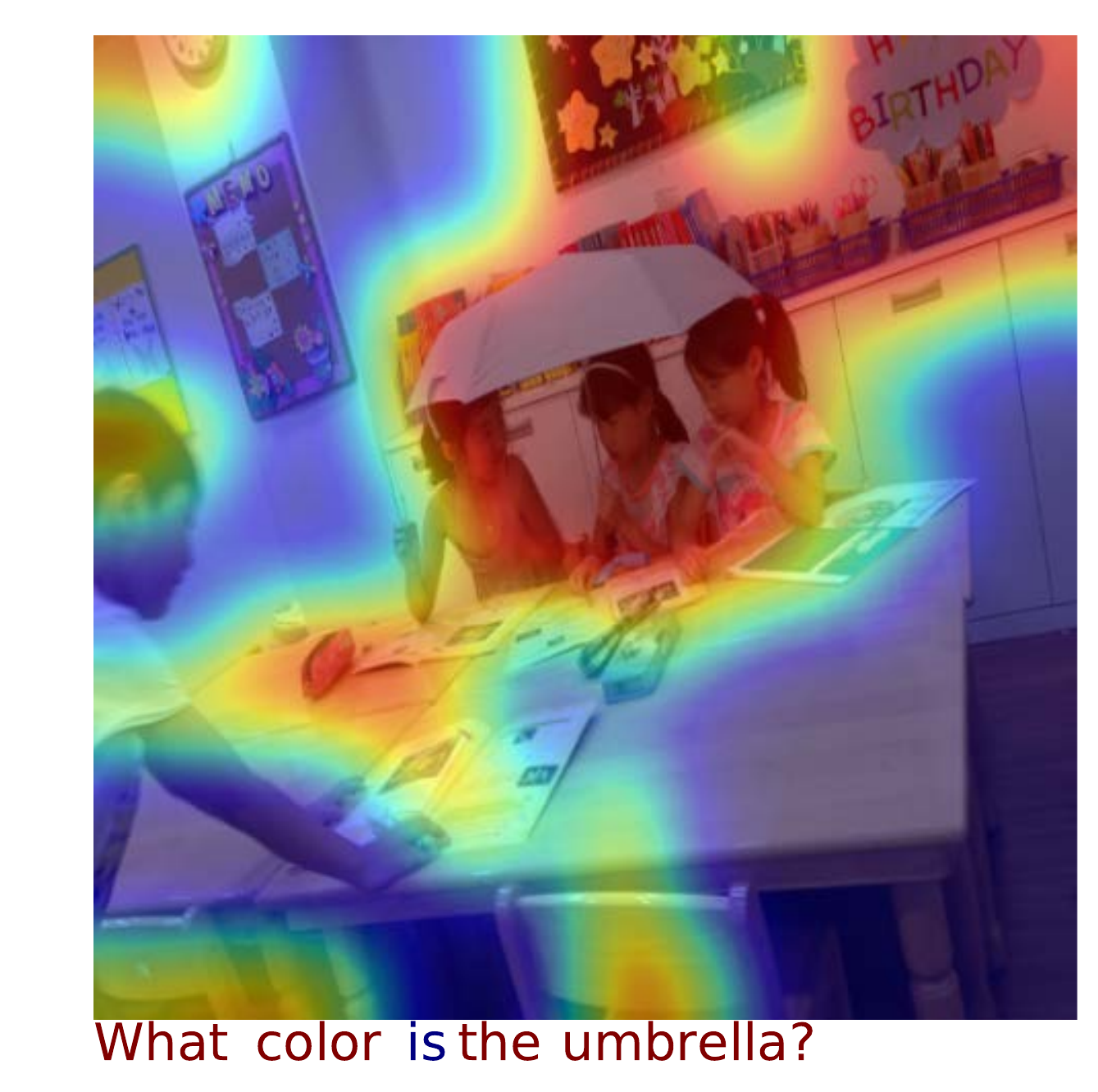} & 		\includegraphics[width=0.175\textwidth,height=65px]{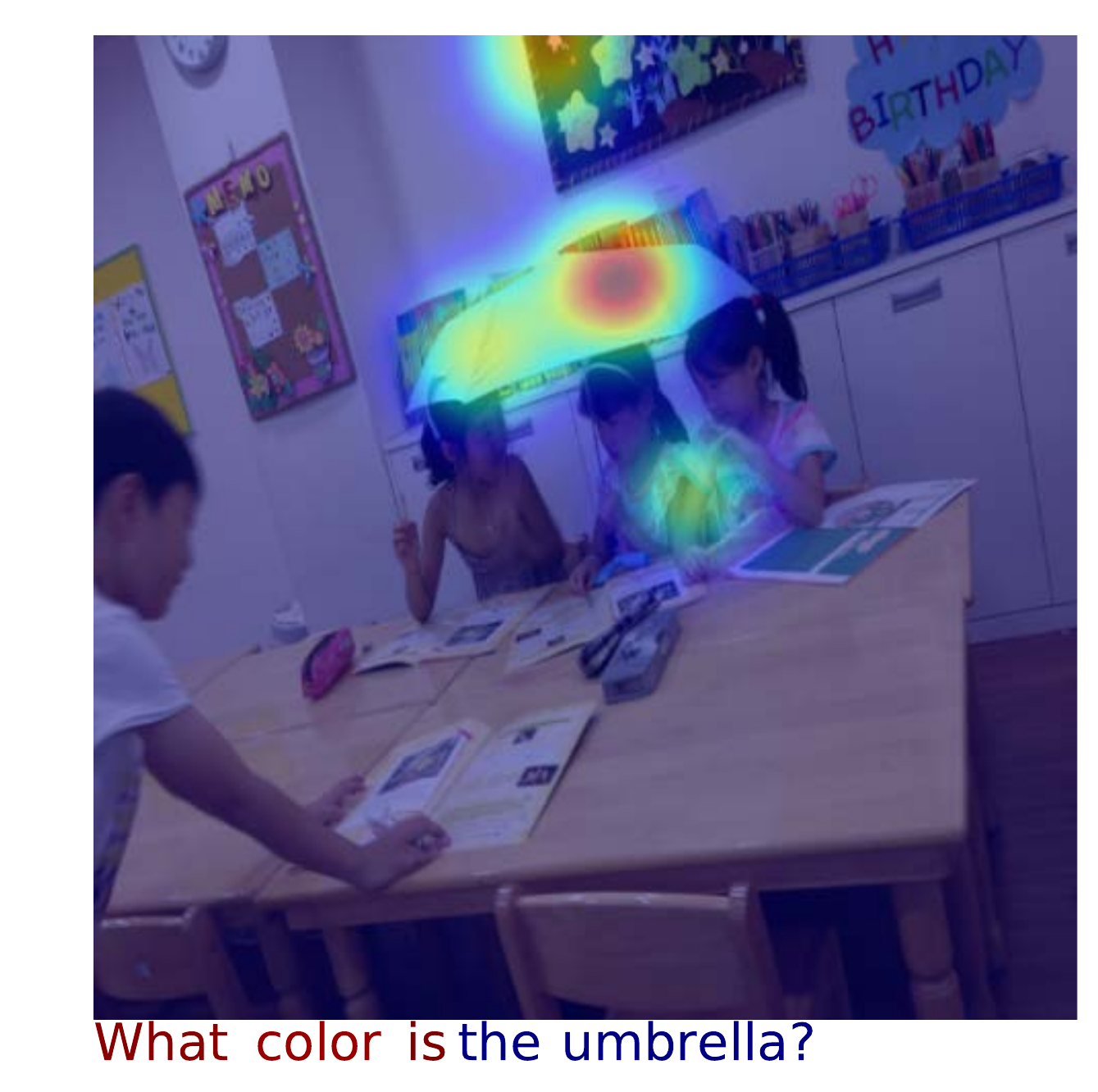} & 		\includegraphics[width=0.175\textwidth,height=65px]{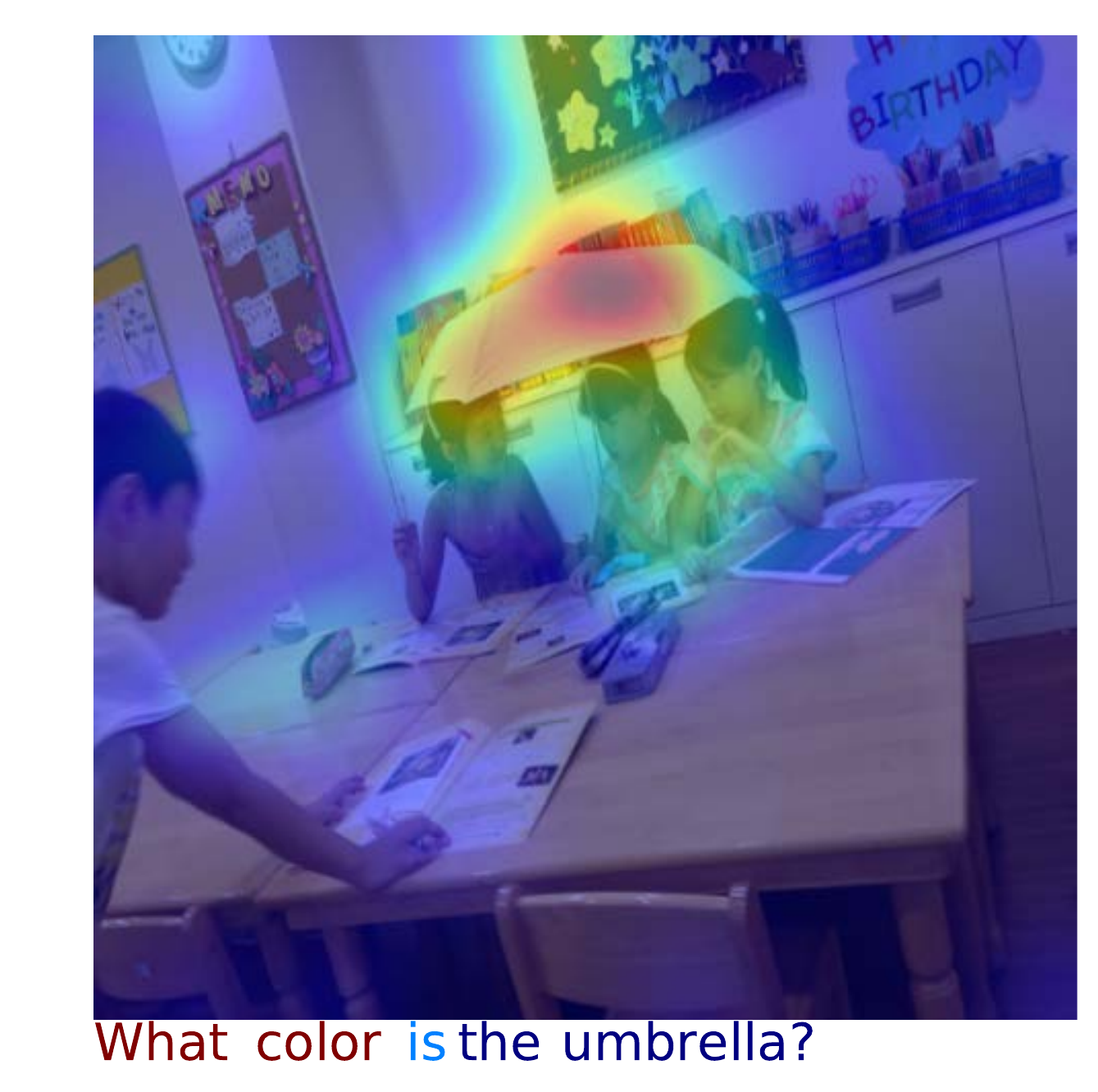}  & 
						\hspace{-0.5cm}\raisebox{1.3cm}{
							{\small{
									\begin{tabular}{l}
										What color is\\ the umbrella? \\
										GT: blue \\
										{\bf Ours:} blue
									\end{tabular}}}}
									
								\end{tabular}
								\vspace{-0.2cm}		
								\caption{
									%{\color{blue}Fill this up with failure case as discussed}
									Failure cases: Unary, pairwise and combined attention of our approach. Our system focuses on the colorful umbrella as opposed to the table in the first row.
								}
								\label{fig:CompFailure}
								\vspace{-0.4cm}
							\end{figure}

In \figref{fig:CompFailure}, we illustrate a failure case, where the attention of our approach is identical, despite two different input questions. Our system focuses on the colorful umbrella as opposed to the object queried for in the question.
	
% !TEX root = nips_2017.tex
\vspace{-0.2cm}
\section{Conclusion}
\vspace{-0.2cm}
In this paper we investigated a series of techniques to design attention  for multimodal input data. Beyond demonstrating state-of-the-art performance using relatively simple models, we hope that this work inspires researchers to work in this direction. %In the future we plan to further extend our models.
% !TEX root = nips_2017.tex
%\vspace{-0.2cm}
%\section*{Acknowledgment}
%\vspace{-0.2cm}

\noindent{\bf Acknowledgments:} This research was supported in part by The Israel Science Foundation (grant No.~948/15). This material is based upon work supported in part by the National Science Foundation under Grant No.~1718221. We thank Nvidia for providing GPUs used in this research.

{\small
	\bibliographystyle{plain}
	\bibliography{egbib}
}

\end{document}